\documentclass{kais}

\usepackage[numbers]{natbib}
\usepackage{amsmath,algpseudocode,float}
\usepackage{amssymb,amsfonts}
\usepackage{algorithm}
\usepackage{algorithmicx}
\usepackage{graphicx}
\usepackage{url}
\newtheorem{defi}{Definition}[section]
\usepackage[justification=centering]{caption}
\usepackage{amsmath}
\usepackage{color} 
\usepackage[misc]{ifsym} 
\usepackage{wrapfig}

\received{04 Jan 2019}
\revised{31 May 2019}
\accepted{04 Aug 2019}

\makeatletter
\newenvironment{breakablealgorithm}
{
	\begin{center}
		\refstepcounter{algorithm}
		\hrule height.8pt depth0pt \kern2pt
		\renewcommand{\caption}[2][\relax]{
			{\raggedright\textbf{\ALG@name~\thealgorithm} ##2\par}%
			\ifx\relax##1\relax 
			\addcontentsline{loa}{algorithm}{\protect\numberline{\thealgorithm}##2}%
			\else 
			\addcontentsline{loa}{algorithm}{\protect\numberline{\thealgorithm}##1}%
			\fi
			\kern2pt\hrule\kern2pt
		}
	}{
		\kern2pt\hrule\relax
	\end{center}
}
\makeatother

\newcommand\blfootnote[1]{%
	\begingroup 
	\renewcommand\thefootnote{}\footnote{#1}%
	\addtocounter{footnote}{-1}%
	\endgroup 
}

\begin{document}
\label{firstpage}

\title{Structured Query Construction via Knowledge Graph Embedding}

\author[R. Wang et al.]{
	Ruijie Wang$^{1,2,7}$, Meng Wang$^{3}$\textsuperscript{(\Letter)}, Jun Liu$^{4,2}$, Michael Cochez$^{5,6,8}$, \and Stefan Decker$^{6,7}$\\
	$^1$National Engineering Lab for Big Data Analytics, Xi'an Jiaotong University, Xi'an, China; \\
	$^2$School of Electronic and Information Engineering, Xi'an Jiaotong University, Xi'an, China;\\
	$^3$School of Computer Science and Engineering, Southeast University, Nanjing, China;\\
	$^4$Guang Dong Xi'an Jiaotong University Academy, Shunde, China;\\
	$^5$VU Amsterdam, Amsterdam, The Netherlands;\\
	$^6$Fraunhofer FIT, Sankt Augustin, Germany;\\
	$^7$Informatik 5, RWTH Aachen University, Aachen, Germany;\\
	$^8$Faculty of Information Technology, University of Jyväskylä, Jyväskylä, Finland.
}

\maketitle

\begin{abstract}
\blfootnote{(\Letter) Meng Wang\\ meng.wang@seu.edu.cn}
In order to facilitate the accesses of general users to knowledge graphs, an increasing effort is being exerted to construct graph-structured queries of given natural language questions. At the core of the construction is to deduce the structure of the target query and determine the vertices/edges which constitute the query. Existing query construction methods rely on question understanding and conventional graph-based algorithms which lead to inefficient and degraded performances facing complex natural language questions over knowledge graphs with large scales. In this paper, we focus on this problem and propose a novel framework standing on recent knowledge graph embedding techniques. Our framework first encodes the underlying knowledge graph into a low-dimensional embedding space by leveraging generalized local knowledge graphs. Given a natural language question, the learned embedding representations of the knowledge graph are utilized to compute the query structure and assemble vertices/edges into the target query. Extensive experiments were conducted on the benchmark dataset, and the results demonstrate that our framework outperforms state-of-the-art baseline models regarding effectiveness and efficiency.
\end{abstract}


\section{Introduction}
\label{Introduction}
\begin{figure*}
	\centering
	\includegraphics[width=0.8\linewidth]{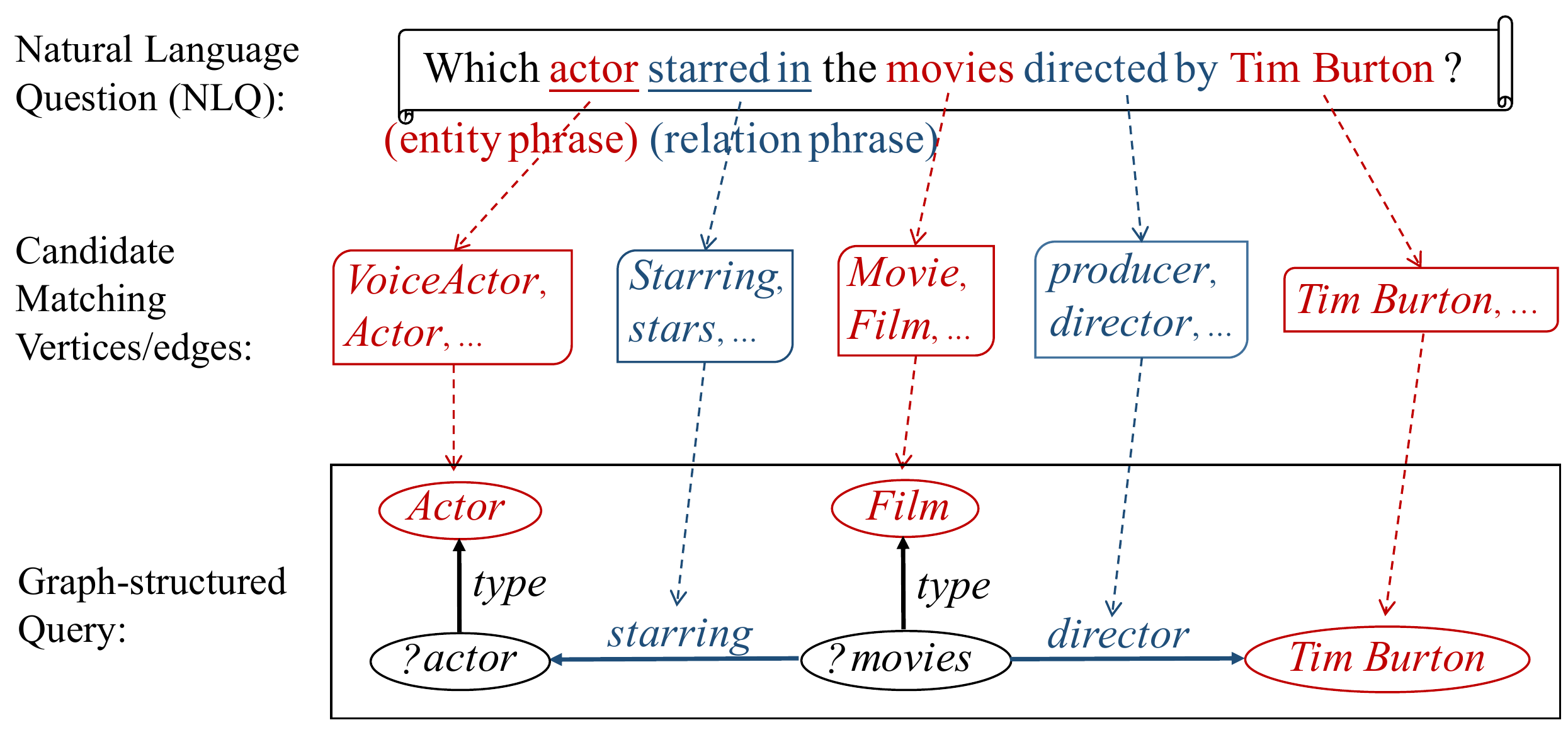}
	\caption{The general query construction process of the example NLQ.}
	\label{newIntroduction}
\end{figure*}

In the past decade, an increasing number of large-scale knowledge graphs (KGs), e.g., DBpedia~\cite{lehmann2015dbpedia} and Wikidata~\cite{vrandevcic2014wikidata}, have been published on the Web. A KG contains a set of triples, e.g., (\textit{Batman}, \textit{director}, \textit{Tim Burton}), each of which consists of two vertices, e.g., \textit{Batman} and \textit{Tim Burton}, and an edge, e.g., \textit{director}. Graph-structured query languages, e.g., SPARQL~\cite{harris2013sparql} and GraphQL~\cite{he2008graphs}, provide an efficient means to retrieve the desired information from KGs. For example, the graph-structured query illustrated in Fig.~\ref{newIntroduction} can be used to retrieve the answer of the question ``\textit{which actor starred in the movies directed by Tim Burton}". Since posing graph-structured queries requires users to be precisely aware of the query syntax and the schema of underlying KGs, general users are more willing to express query intentions with natural language questions (NLQs). To hide the complexity of query languages, numerous models~\cite{zheng2017natural,yin2015answering,yahya2012natural,hu2018answering,zou2014natural} have been proposed to construct graph-structured queries of given NLQs.

\begin{figure*}
	\centering
	\includegraphics[width=0.5\linewidth]{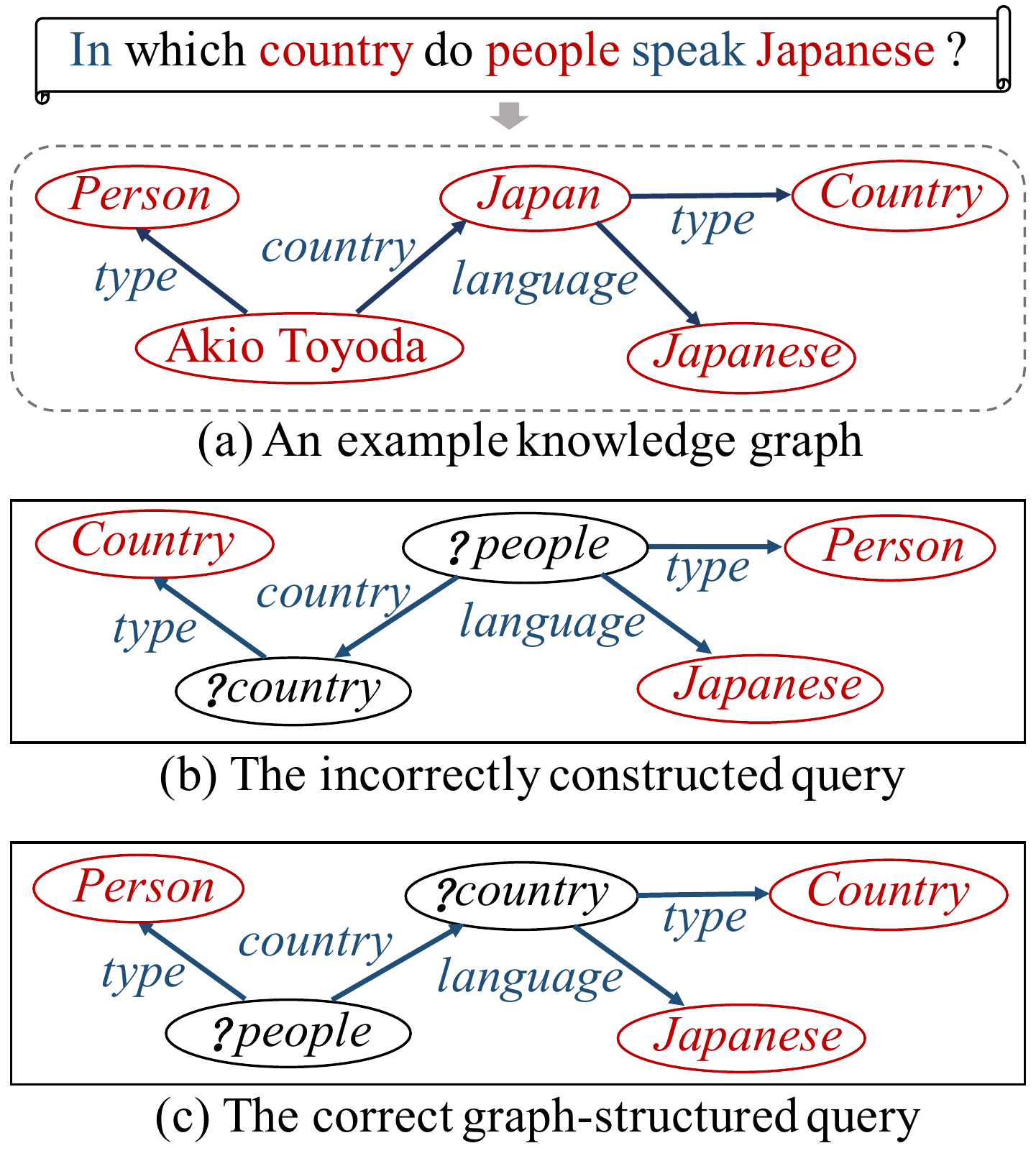}
	\caption{An example of the \textit{semantic gap} between NLQs and KGs.}
	\label{japanese}
\end{figure*}

\textbf{Challenges:} A widely adopted pipeline of the query construction mainly includes two phases, as illustrated in Fig.~\ref{newIntroduction}. The first phase is to map entity/relation phrases of the NLQ to their matching vertices/edges in the underlying KG. The second phase is to assemble matching vertices/edges into the target graph-structured query according to the deduced query structure. We need to address the following challenges during the construction:

\begin{enumerate}

\item An entity/relation phrase may have multiple candidate matching vertices/edges in the underlying KG, and it is hard to select the most suitable one. Taking the entity phrase ``actor" as an example, it can be mapped to multiple candidate vertices including \textit{Actor}, \textit{Artist}, and \textit{VoiceActor}. With candidate matching vertices/edges, there exist models~\cite{zou2014natural,hu2018answering} which first construct a set of candidate queries and then verify them over KGs, which are inefficient facing KGs with large scales. Other models~\cite{yahya2012natural,zheng2017natural} try to prune the phrase mapping results before the query generation, but they may filter out the optimal candidates.

\item Existing query construction models~\cite{yin2015answering,yahya2012natural,hu2018answering,zou2014natural} rely on question understanding to deduce the target query structure without considering the underlying KG. These models cannot handle the ``\textit{semantic gap}"\footnote{The ``\textit{semantic gap}" refers to that KGs organize structured information differently from what one can deduce from natural language expressions~\cite{diefenbach2018core}.} between NLQs and KGs. Let us consider another NLQ ``\textit{In which country do people speak Japanese}" which is posed over the KG illustrated in Fig.~\ref{japanese}(a). From the perspective of question understanding, there is an obvious semantic relation ``speak" between ``people" and ``Japanese", and the query shown in Fig.~\ref{japanese}(b) is very likely to be constructed. However, in the underlying KG, people are not linked to languages, and the correct query is shown in Fig.~\ref{japanese}(c).

\end{enumerate}

\textbf{Our Solution:} In this paper, we focus on the above challenges and propose a novel graph embedding-based framework to construct graph-structured queries of NLQs. Our framework contains the following processes:

Firstly, in the offline stage, the underlying KG is encoded into a low-dimensional embedding space based on \textit{generalized local knowledge graphs} which represent the contexts of vertices/edges. Then, in the online stage, each entity/relation phrase of the given NLQ is mapped to a set of candidate matching vertices/edges of the underlying KG. With the embedding vectors learned in the offline stage, the mapping results are utilized to compute the structure of the target query. Finally, we select the most suitable matching vertices/edges and assemble them into the target query according to the computed query structure.

\textbf{Contributions:} In a nutshell, our work makes the following contributions:

\begin{enumerate}
	\item We propose a novel graph embedding-based framework to construct graph-structured queries of NLQs.
	
	\item We propose a translation-based embedding method which leverages the \textit{generalized local knowledge graphs} to make the learned embedding vectors applicable to the query construction task.
	
	\item We propose effective and efficient approaches to compute the structure of the target query and determine the most suitable matching vertices/edges based on the learned embedding vectors.
	
	\item We conducted extensive experiments on the benchmark dataset to evaluate our framework. The results show that our method outperforms several state-of-the-art baselines regarding both effectiveness and efficiency.
\end{enumerate}

\textbf{Organization:} The remainder of this paper is organized as follows: Section~\ref{Background} introduces the background of this paper. Section~\ref{Proposed_Framework} presents our proposed framework in detail. The evaluation of the framework is reported in Section~\ref{Experiments}. Related work is discussed in Section~\ref{Related_Work}. Finally, conclusions and future work are presented in Section~\ref{Conclusion}.

\section{Background}
\label{Background}
For a broader view of the graph-structured query construction task, we briefly introduce the knowledge graph (KG) and KG embedding techniques in this section.

\subsection{Knowledge Graph}
A KG is an integration of the extensive real-world information, which is organized as a labeled directed graph, where vertices represent entities, and directed edges represent semantic relations between entities. Here we present the notations of the KG used in this paper as follows: Let $\mathcal{V}$ be a set of vertices (e.g., \textit{Batman}, and \textit{Tim Burton}), $\mathcal{E}$ be a set of edges (e.g., \textit{director}). A KG triple, e.g., (\textit{Batman}, \textit{director}, \textit{Tim Burton}), is denoted as $(v_h, e, v_t)$, where $v_h, v_t\in \mathcal{V}$ and $e\in \mathcal{E}$. As a finite set of KG triples, the KG is denoted as $\mathcal{G} = (\mathcal{V}, \mathcal{E})$. 

Following the Semantic Web standards, e.g., RDF~\cite{klyne2006resource} and OWL~\cite{antoniou2004web}, KGs are unified, interchangeable, disambiguated, and have reasoning capabilities. As a large-scale integration of the real-world information with these features, the KG can significantly facilitate knowledge-based tasks including question answering~\cite{hu2018answering,zou2014natural}, information retrieval~\cite{corcoglioniti2016knowledge}, natural language processing (NLP)~\cite{schuhmacher2014knowledge}, and recommending~\cite{palumbo2017entity2rec}. In the past decade, KGs have gained considerable attention in both academia and industry. For example, DBpedia was initially released in 2007, since then DBpedia has been widely adopted as the benchmark dataset of question answering challenges (e.g., QALD\footnote{\url{http://qald.aksw.org/}}). And the latest version of DBpedia contains NLP Interchange Format (NIF)~\cite{hellmann2012nif} annotations which can facilitate NLP tasks. In 2012, Google announced the Google Knowledge Graph\footnote{\url{https://googleblog.blogspot.com/2012/05/introducing-knowledge-graph-things-not.html}}, which enables its search engine to search for "things" rather than "strings".

With the rapid growth of KGs, some non-trivial challenges have arisen. Firstly, KGs often suffer from the incompleteness, sparseness, and noise issues since most of them are built either collaboratively or semi-automatically~\cite{xu2016knowledge}. KG refinement approaches, including completion and correction models~\cite{lin2015learning,shi2018open,zhao2019scef}, have been proposed for these issues. Secondly, due to the graph structure of KGs and the common adoption of RDF standard, graph-structured queries, e.g., SPARQL and GraphQL, are recognized as basic facilities for accessing KGs. However, graph-structured queries are too technical for general users, as we analyzed in Section~\ref{Introduction}, which is another challenge and also the motivation of this paper. Thirdly, due to the large scales of existing KGs, performances of conventional graph-based algorithms over KGs are compromised by data sparsity and computational inefficiency issues. Recently, KG embedding techniques~\cite{bordes2013translating,wang2014knowledge,feng2016gake,shi2017knowledge} have been proposed to address this challenge.

\subsection{KG Embedding Techniques}
 KG embedding techniques learn the vectorized representations of KGs in low-dimensional vector spaces, where vertices and edges are represented by embedding vectors, and the essential information of KGs, e.g., structural relations among vertices and edges, is modeled by specifically designed mechanisms.
 
 \begin{figure*}
 	\centering
 	\includegraphics[width=0.65\linewidth]{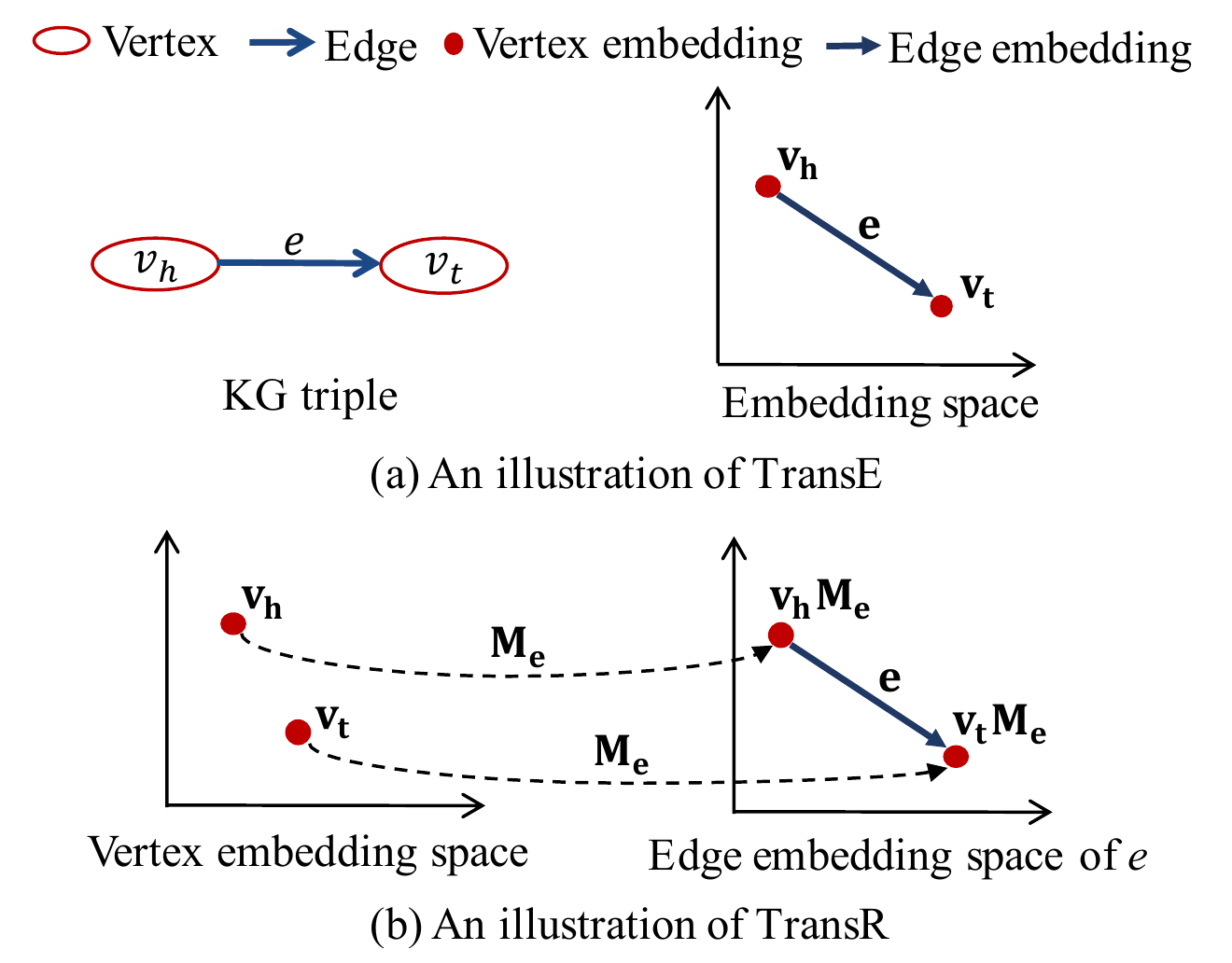}
 	\caption{Illustrations of the translation mechanisms of TransE and TransR.}
 	\label{TransE}
 \end{figure*}

A mainstream of KG embedding is the translation-based models~\cite{feng2016gake}, including TransE~\cite{bordes2013translating} and its variants~\cite{wang2014knowledge,lin2015learning,lin2015modeling}, which represent edges as translation operations from head vertices to tail vertices in embedding spaces, as illustrated in Fig.~\ref{TransE}(a). Specifically, given a KG $\mathcal{G} = (\mathcal{V}, \mathcal{E})$, for two vertices $v_h, v_t\in \mathcal{V}$, and an edge $e\in \mathcal{E}$, we use boldface letters $\mathbf{v_h}$, $\mathbf{v_t}$, and $\mathbf{e}$ to denote their embedding vectors. If $(v_h, e, v_t) \in \mathcal{G}$, the translation mechanism of TransE requires $\mathbf{v_h} + \mathbf{e} \approx \mathbf{v_t}$, i.e., $\mathbf{v_t}$ should be the closest neighbor of $\mathbf{v_h} + \mathbf{e}$ in the embedding space. Inspired by TransE, improved models, e.g., TransR~\cite{wang2014knowledge}, TransH~\cite{lin2015learning}, and PTransE~\cite{lin2015modeling}, were later proposed to achieve better performances. Taking TransR as an example, it encodes KG into an vertex embedding space and multiple edge embedding spaces. For vertices $v_h, v_t\in \mathcal{V}$, and the edge $e\in \mathcal{E}$, besides embedding vectors $\mathbf{v_h}$, $\mathbf{v_t}$, and $\mathbf{e}$, TransR also learns the projection matrix $\mathbf{M_e}$ which projects vertices from the vertex embedding space to the edge embedding space specified by $e$, as illustrated in Fig.~\ref{TransE}(b). Formally, if $(v_h, e, v_t)\in \mathcal{G}$, TransR requires that $\mathbf{v_h}\mathbf{M_e} + \mathbf{e} \approx \mathbf{v_t}\mathbf{M_e}$. Translation-based models capture the structural information of KGs precisely, and they have been adopted in tasks such as KG completion\cite{lin2015learning} and graph-structured query construction\cite{han2017keyword}. However, translation-based models ignore contextual information and are not suitable for tasks such as classification and regression~\cite{ristoski2016rdf2vec}.

Context-based models~\cite{wang2018towards,feng2016gake,ristoski2016rdf2vec,shi2017knowledge} which consider the contextual information of KGs have been proposed in recent years. GAKE~\cite{feng2016gake} defines three kinds of context for vertices and edges, including neighbor context, edge context, and path context. During learning, GAKE maximizes the conditional probability of each vertex/edge given its context. Therefore, embedding representations learned by GAKE are able to predict missing vertices and edges given their contexts. RDF2Vec~\cite{ristoski2016rdf2vec} does not define the context of KGs. It transforms KGs into sequences of vertices and encodes vertices by neural language models, i.e., Continuous Bag-of-Words model (CBOW) and Skip-Gram model. However, we still regard RDF2Vec as a context-based model since the neural language models are trained based on context windows of the sequences which can be considered as contexts of vertices.

\section{Proposed Framework}
\label{Proposed_Framework}

In this section, we first introduce the notations employed in this paper and then elaborate on our framework.

\textbf{Entity Vertex and Class Vertex.} We divide vertices of KGs into two categories: the entity vertex $v_e\in \mathcal{V}$ representing a specific entity, and the class vertex ${v_c}\in \mathcal{V}$ representing a class of entity vertices. We deduce the categories of vertices according to the type-related statements of KGs. For example, according to the KG triple (\textit{Batman}, \textit{type}, \textit{Film}), \textit{Batman} is an entity vertex, and its class vertex is \textit{Film}.

\textbf{NLQ and Entity/Relation Phrase.} We denote the natural language question (NLQ) as $\mathcal{Q}$. The entity phrase (e.g., ``actor", ``movies", and ``Tim Burton") and the relation phrase (e.g., ``starred in" and ``directed by") in $\mathcal{Q}$ are denoted as $ent$ and $rel$, respectively.

\textbf{Graph-Structured Query.} Let $\mathcal{V}_v$ be a set of variables\footnote{In this paper, we focus on the NLQs whose answers are vertices in the underlying KG. Therefore, only vertex variables will be considered.}, where the variable $v_v\in \mathcal{V}_v$ is distinguished from vertices by a leading question mark symbol, e.g., \textit{?movies}. A triple pattern is similar to the KG triple but allows the use of variables, e.g., (\textit{?movies}, \textit{director}, \textit{Tim Burton}). We define the graph-structured query $\mathcal{G_Q}$ as a finite set of triple patterns.

\begin{figure*}
	\centering
	\includegraphics[width=0.9\linewidth]{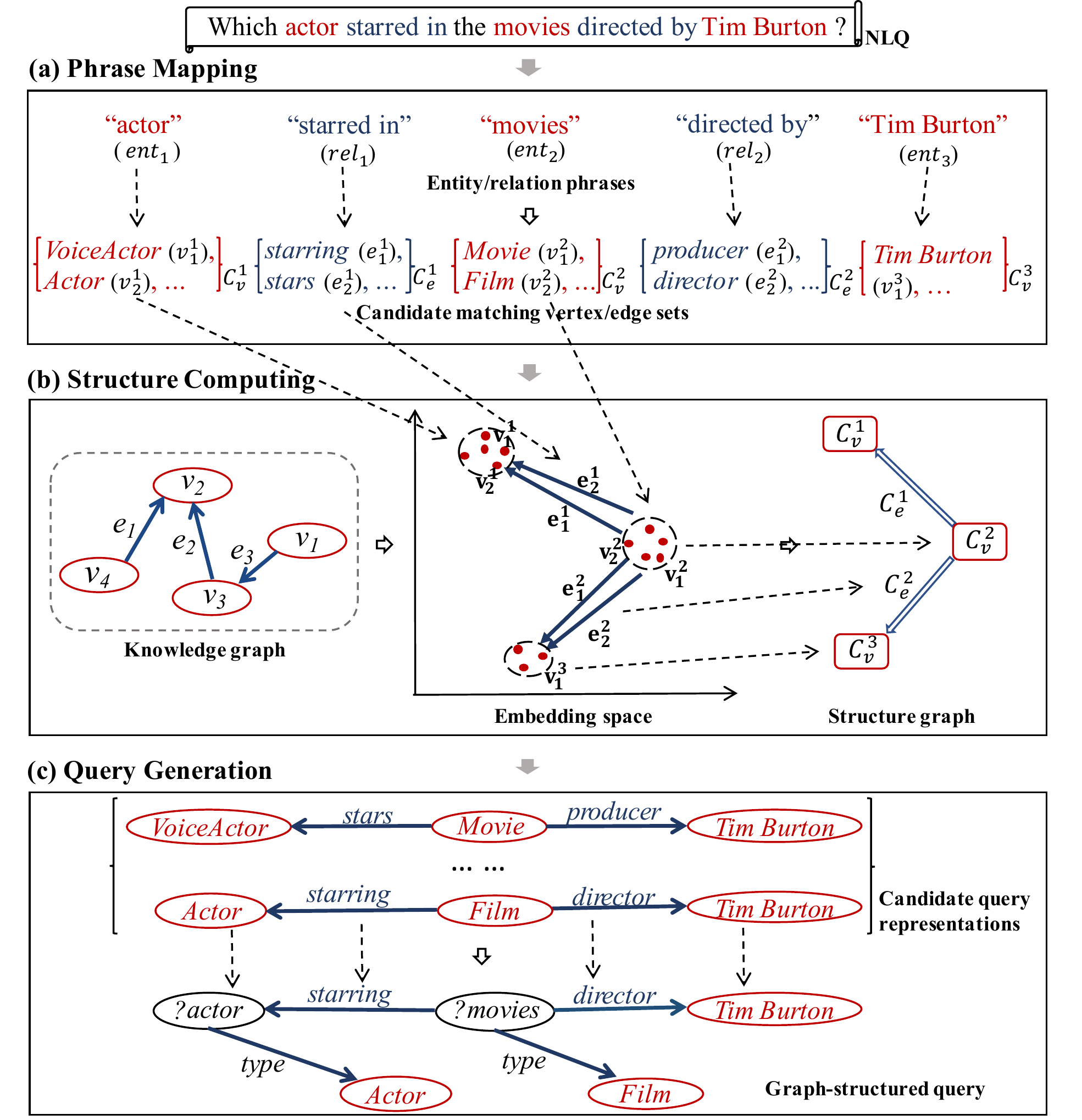}
	\caption{An overview of our framework.}
	\label{framework}
\end{figure*}

\subsection{Overview of Proposed Framework}
\label{Overview of Proposed Framework}
Our framework constructs graph-structured queries of given NLQs through three modules: phrase mapping, structure computing, and query generation. We depict an overview of the proposed framework in Fig.~\ref{framework}.

In the first module, each entity/relation phrase of the NLQ is mapped to a set of candidate matching vertices/edges. We denote the candidate vertex set and the candidate edge set as $C_v$ and $C_e$, respectively. For example, the candidate vertex set of the entity phrase ``actor" ($ent_1$) is $C_v^1=\{v_1^1,v_2^1,...\}$ (i.e., \{\textit{VoiceActor}, \textit{Actor}, ...\}), as illustrated in Fig.~\ref{framework}(a).

In the second module, embedding vectors learned in the offline stage are utilized to compute the structure of the target query, as shown in Fig.~\ref{framework}(b). Note that we require vertices/edges in the same candidate set should be close to each other in the embedding space. For example, $\mathbf{v_1^1}$ should be close to $\mathbf{v_2^1}$ since \textit{VoiceActor} ($v_1^1$) and \textit{Actor} ($v_2^1$) are in the same candidate set. Then, each candidate vertex/edge set can be represented by a mean embedding vector, and we adopt the translation mechanism of TransE to compute the target query structure which is represented by the structure graph consisting of candidate vertex/edge sets. For example, the target query structure of the example NLQ is $\{(C_v^2, C_e^1, C_v^1),(C_v^2, C_e^2, C_v^3)\}$, as shown on the right of Fig.~\ref{framework}(b).

In the third module, we assemble candidate matching vertices/edges into a set of candidate query representations and evaluate them to generate the target query, as illustrated in Fig.~\ref{framework}(c). The evaluation is also based on the learned embedding vectors. For instance, the generated query of the example NLQ is \{(\textit{?movies}, \textit{director}, \textit{Tim Burton}), (\textit{?movies}, \textit{starring}, \textit{?actor}), (\textit{?actor}, \textit{type}, \textit{Actor}), (\textit{?movies}, \textit{type}, \textit{Film})\}.

\subsection{KG Embedding Learning}
\label{KG Embedding Learning}

As introduced above, we have the following requirements for the learned embedding vectors. Firstly, vertices/edges in the same candidate set should be close to each other in the embedding space. Secondly, the translation mechanism should be maintained for the structure computing. In addition, we represent variables (e.g., \textit{?movies} and \textit{?actor}) of the target query by their class vertices (e.g., \textit{Film} and \textit{Actor}) in the candidate query representations, as shown in Fig.~\ref{framework}(c). Since the candidate query representations are evaluated based on the learned embedding vectors, the embedding learning method is also required to be able to capture the relations relevant to class vertices, including the relation between two class vertices, e.g., (\textit{Film}, \textit{starring}, \textit{Actor}), and the relation between a class vertex and an entity vertex, e.g., (\textit{Film}, \textit{starring}, \textit{Tim Burton}). 

Translation-based models satisfy the requirement of the translation mechanism. However, they do not consider the semantics of vertices/edges, and the first requirement would not be satisfied. Context-based models utilize the context information to represent vertices/edges in the embedding space. Since the vertices/edges of the same disambiguated candidate set share common context information, the first requirement can be satisfied by context-based models. However, most context-based models do not maintain the translation mechanism. Besides that, in the underlying KG, except type-related statements, e.g., (\textit{Batman}, \textit{type}, \textit{Film}), the relations relevant to class vertices are rarely described. None of existing translation/context-based models considers this issue, and they do not satisfy the third requirement. 

In this section, we propose a novel embedding method which leverages \textit{generalized local knowledge graphs} (GL-KGs) to learn required embedding vectors. For each vertex/edge of the underlying KG, its GL-KG is constructed by generalizing all the triples relevant to the vertex/edge. For example, given KG triples (\textit{Batman}, \textit{starring}, \textit{Michael Keaton}), (\textit{Batman}, \textit{type}, \textit{Film}), and (\textit{Michael Keaton}, \textit{type}, \textit{Actor}), we can deduce the generalized relation between \textit{Film} and \textit{Actor} as (\textit{Film}, \textit{starring}, \textit{Actor}). Based on the \textit{local knowledge graph} (L-KG), we formally define the GL-KG as follows:

\begin{defi} [Local Knowledge Graph] \label{Local Knowledge Graph}
	Given a KG $\mathcal{G} = (\mathcal{V}, \mathcal{E})$, the local knowledge graph (L-KG) of the vertex $v\in \mathcal{V}$ is the KG triple set $\mathcal{G_L}(v) = \{(v, e, \hat{v}) | e\in \mathcal{E}, \hat{v}\in \mathcal{V}, (v, e, \hat{v})\in \mathcal{G}\} \cup \{(\hat{v}, e, v) | \hat{v}\in \mathcal{V}, e\in \mathcal{E}, (\hat{v}, e, v)\in \mathcal{G}\}$. The L-KG of the edge $e\in \mathcal{E}$ is the KG triple set $\mathcal{G_L}(e) = \{(v, e, \hat{v}) | v, \hat{v}\in \mathcal{V}, (v, e, \hat{v})\in \mathcal{G}\}$. 
\end{defi}

\begin{defi} [Generalized Local Knowledge Graph] \label{Generalized Local Knowledge Graph} Given a KG $\mathcal{G} = (\mathcal{V}, \mathcal{E})$, the generalized local knowledge graph (GL-KG) of entity vertex $v_e\in \mathcal{V}$ is the triple set $\mathcal{G_G}(v_e) = \{(v_e, e, \hat{v_c})|(v_e, e, \hat{v_e})\in \mathcal{G_L}(v_e)\}\cup \{(\hat{v_c}, e, v_e)|(\hat{v_e}, e, v_e)\in \mathcal{G_L}(v_e)\}$, where $\hat{v_c}$ is the class vertex of $\hat{v_e}$. The GL-KG of class vertex $v_c\in \mathcal{V}$ is the triple set $\mathcal{G_G}(v_c) = \{(v_c, e, \hat{v_c})|(v_e, e, \hat{v_c})\in \mathcal{G_G}(v_e)\}\cup \{(\hat{v_c}, e, v_c)|(\hat{v_c}, e, v_e)\in \mathcal{G_G}(v_e)\}$, where $v_c$ is the class vertex of $v_e$. The GL-KG of edge $e\in \mathcal{E}$ is the triple set $\mathcal{G_G}(e) = \{(v_c, e, \hat{v_c}),(v_e, e, \hat{v_c}),(v_c, e, \hat{v_e})|(v_e, e, \hat{v_e})\in \mathcal{G_L}(e)\}$, where $v_c$ and $\hat{v_c}$ are class vertices of $v_e$ and $\hat{v_e}$, respectively.
\end{defi}

\begin{figure*}
	\centering
	\includegraphics[width=0.95\linewidth]{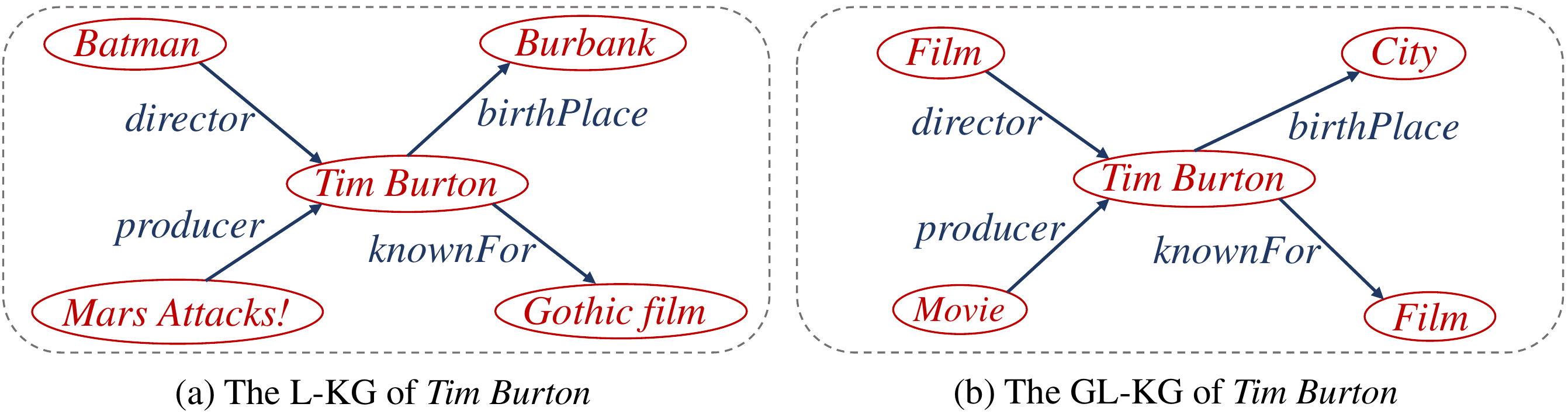}	
	\caption{The L-KG and GL-KG of the entity vertex \textit{Tim Burton}.}
	\label{generalizedGraph}
\end{figure*}

Taking the entity vertex \textit{Tim Burton} as an example, its L-KG is illustrated in Fig.~\ref{generalizedGraph}(a), and its GL-KG is illustrated in Fig.~\ref{generalizedGraph}(b).

Our objective is to learn embedding vectors of vertices/edges according to their GL-KGs. To this end, we first define the conditional probability of vertex $v\in \mathcal{V}$ given its GL-KG $\mathcal{G_G}(v)$ as follows:
\begin{equation}
\label{vertex softmax}
P(v|\mathcal{G_G}(v))=\frac{\exp(f_1(v,\mathcal{G_G}(v)))}{\sum_{v'\in \mathcal{V}}\exp(f_1(v',\mathcal{G_G}(v)))},
\end{equation}
where $f_1(v',\mathcal{G_G}(v))$ is the function that measures the correlation between an arbitrary vertex $v'\in \mathcal{V}$ and $\mathcal{G_G}(v)$. The above equation can be considered as the compatibility between the vertex $v$ and its GL-KG, and it is formulated as a softmax-like representation which has been validated~\cite{shi2017knowledge}.

Then, we adopt the translation mechanism of TransE to define the function $f_1(v',\mathcal{G_G}(v))$ as follows:
\begin{equation}
\label{vertex score function}
\begin{aligned}
f_1(v',\mathcal{G_G}(v)) = &-\frac{1}{A(\mathcal{G_G}(v))}(\sum_{(v,e,\hat{v_c})\in \mathcal{G_G}(v)} a(v,e,\hat{v_c})\cdot\| \mathbf{v'} + \mathbf{e} - \mathbf{\hat{v_c}} \|^2_2 \\&+ \sum_{(\hat{v_c},e,v)\in \mathcal{G_G}(v)} a(\hat{v_c},e,v)\cdot\| \mathbf{\hat{v_c}} + \mathbf{e} - \mathbf{v'} \|^2_2),
\end{aligned}
\end{equation}
\begin{equation}
A(\mathcal{G_G}(v)) = \sum_{(v,e,\hat{v_c})\in \mathcal{G_G}(v)}a(v,e,\hat{v_c})+\sum_{(\hat{v_c},e,v)\in \mathcal{G_G}(v)}a(\hat{v_c},e,v),
\end{equation}
where $a(v,e,\hat{v_c})$ is the attention score of the triple $(v,e,\hat{v_c})$, which is computed by the following equation:
\begin{equation}
	\label{attention-score-entity}
a(v,e,\hat{v_c}) = \exp(\frac{|\{\hat{v_e}|(v,e,\hat{v_e})\in \mathcal{G_L}(v) \}|}{|\mathcal{G_L}(v)|}),
\end{equation}
where $\hat{v_e}$ is the entity vertex of $\hat{v_c}$, and $|\mathcal{G_L}(v)|$ denotes the size of $\mathcal{G_L}(v)$. $a(\hat{v_c},e,v)$ is computed analogically.

The intuition of the attention score is that when encoding the vertex $v$, different triples in its GL-KG $\mathcal{G_G}(v)$ may attract different attention. If a triple in $\mathcal{G_G}(v)$ can be generalized from more KG triples in its L-KG $\mathcal{G_L}(v)$, this triple should have more impacts. For example, \textit{Tim Burton} is a director and a producer at the same time, and two triples (\textit{Movie}, \textit{director}, \textit{Tim Burton}) and (\textit{Movie}, \textit{producer}, \textit{Tim Burton}) exist in its GL-KG. Since (\textit{Movie}, \textit{director}, \textit{Tim Burton}) can be generalized by more KG triples in the L-KG of \textit{Tim Burton}, i.e., the number of movies directed by \textit{Tim Burton} is larger than the number of movies produced by \textit{Tim Burton}, the triple (\textit{Movie}, \textit{director}, \textit{Tim Burton}) should attract more attention when encoding \textit{Tim Burton}. 

Embedding vectors of vertices can be learned by maximizing the joint probability of all vertices in $\mathcal{V}$, which is formulated as follows:
\begin{equation}
\label{vertex objective}
O_{v}=\sum_{v\in \mathcal{V}}\log P(v|\mathcal{G_G}(v)).
\end{equation}

Analogically, we define the conditional probability of $e\in \mathcal{E}$ given its GL-KG $\mathcal{G_G}(e)$ as follows:
\begin{equation}
\label{edge softmax}
P(e|\mathcal{G_G}(e))=\frac{\exp(f_2(e,\mathcal{G_G}(e)))}{\sum_{e'\in \mathcal{E}}\exp(f_2(e',\mathcal{G_G}(e)))},
\end{equation}
where $f_2(e',\mathcal{G_G}(e))$ is the function that measures the correlation between an arbitrary edge $e'\in \mathcal{E}$ and $\mathcal{G_G}(e)$. $f_2(e',\mathcal{G_G}(e))$ is also formulated based on the translation mechanism:
\begin{equation}
\label{edge score function}
f_2(e',\mathcal{G_G}(e))=-\frac{1}{\sum_{(v,e,\hat{v})\in \mathcal{G_G}(e)} a'(v,e,\hat{v})}(\sum_{(v,e,\hat{v})\in \mathcal{G_G}(e)} a'(v,e,\hat{v})\cdot\|\mathbf{v}+\mathbf{e'}-\mathbf{\hat{v}}\|^2_2),
\end{equation}
where $a'(v, e, \hat{v})$ is the function that computes the attention score of the triple $(v, e, \hat{v})$, defined as follows:
\begin{equation}
a'(v, e, \hat{v})=\left\{
\begin{aligned}
\exp(\frac{|\{v_e|(v_e, e, \hat{v})\in\mathcal{G_L}(e)\}|}{|\mathcal{G_L}(e)|}),
& \text{ iff } \hat{v} \text{ is an entity vertex},\\
\exp(\frac{|\{\hat{v_e}|(v, e, \hat{v_e})\in\mathcal{G_L}(e)\}|}{|\mathcal{G_L}(e)|}),
& \text{ iff } v \text{ is an entity vertex},\\
\exp(\frac{|\{(v_e, \hat{v_e})|(v_e, e, \hat{v_e})\in\mathcal{G_L}(e)\}|}{|\mathcal{G_L}(e)|}),
& \text{ if } v \text{ and } \hat{v} \text{ are class vertices},
\end{aligned}
\right.
\end{equation}
where $v_e$ and $\hat{v_e}$ are respectively entity vertices of $v$ and $\hat{v}$ when $v$ and $\hat{v}$ are class vertices.

Embedding vectors of edges can be learned by maximizing the joint probability of all edges in $\mathcal{E}$, formulated as follows:
\begin{equation}
\label{edge objective}
O_e=\sum_{e\in \mathcal{E}}\log P(e|\mathcal{G_G}(e)).
\end{equation}

We jointly maximize the objective functions of vertices and edges to learn the required embedding vectors:
\begin{equation}
\label{joint objective}
O=\lambda_v O_v + \lambda_e O_e,
\end{equation}
where $\lambda_v$ and $\lambda_e$ are weighting hyper-parameters.

It is impractical to directly compute Equ.~\ref{vertex softmax} and Equ.~\ref{edge softmax} due to the large scale of the underlying KG. Therefore, we follow~\cite{mikolov2013distributed} to approximate them based on negative sampling. Taking Equ.~\ref{vertex softmax} as an example, it can be approximated by the following equation:
\begin{equation}
\label{approximate vertex}
P(v|\mathcal{G_G}(v))\approx \sigma(f_1(v,\mathcal{G_G}(v)))\cdot \prod_{v'\in \mathcal{V}^v_\mathcal{N}}^{n} \sigma(-f_1(v',\mathcal{G_G}(v))),
\end{equation}
where $n$ is the number of negative samples, $\sigma(\cdot)$ is the sigmoid function, and $v'$ is the negative vertex which is obtained by sampling vertices from a uniform distribution over the negative vertex set $\mathcal{V}^v_\mathcal{N}$. For each negative vertex $v'\in \mathcal{V}^v_\mathcal{N}$, we require that $\mathcal{G_G}(v')\cap \mathcal{G_G}(v) = \varnothing$.

Intuitively, vertices/edges in the same disambiguated candidate set usually share common GL-KGs. For example, \textit{Actor} and \textit{VoiceActor} are both linked to films with high attention scores, and \textit{starring} and \textit{stars} link actor-film pairs with high attention scores. According to Equ.~\ref{vertex softmax} and Equ.~\ref{edge softmax}, the learned embedding vectors of vertices/edges which have similar GL-KGs would be close to each other. Therefore, the first requirement is satisfied. Our embedding method also maintains the translation mechanism of TransE since the translation mechanism is adopted in Equ.~\ref{vertex score function} and Equ.~\ref{edge score function}. In addition, embedding vectors are learned based on GL-KGs which contain generalized KG triples, e.g., (\textit{Film}, \textit{starring}, \textit{Actor}), our embedding method is able to capture the relations relevant to class vertices. It is worth mentioning that the learned embedding vectors can be utilized in the following online query construction modules without any further modification.

\subsection{Phrase Mapping}
\label{Phrase Mapping}
In this module, we extract entity/relation phrases from the given NLQ and map each phrase to a set of candidate matching vertices/edges, as illustrated in Fig.~\ref{framework}(a). Since the extraction and mapping are not the focus of this paper, and they have been well studied in previous works~\cite{hu2018answering,yahya2012natural,dubey2018earl,mihalcea2007wikify}, we adopt the existing methods~\cite{dubey2018earl,hu2018answering} to obtain candidate vertex/edge sets.

Following~\cite{dubey2018earl}, we first use SENNA~\cite{collobert2011natural} to extract keywords from the given NLQ. Then, a character embedding based long short-term memory network (LSTM)~\cite{dubey2018earl} is trained to classify the extracted keywords into entity phrases and relation phrases. We denote entity phrases as $\{ent_1, ent_2, ..., ent_n\}$ and relation phrases as $\{rel_1, rel_2, ..., rel_m\}$. For example, given the NLQ ``\textit{which actor starred in the movies directed by Tim Burton}", we expect to obtain entity phrases \{``actor", ``movies", ``Tim Burton"\} and relation phrases \{``starred in", ``directed by"\}. Then, an exhaustive list of candidate matching vertices/edges is retrieved for each entity/relation phrase by querying an Elasticsearch\footnote{\url{https://www.elastic.co/}} index of vertex/edge-label pairs. Considering semantic equivalence and grammatical variations, Dubey et al.~\cite{dubey2018earl} created the index based on Wikidata labels, Oxford Dictionary API\footnote{\url{https://developer.oxforddictionaries.com/}}, and fastText\footnote{\url{https://fasttext.cc/}}. For example, the exhaustive list of ``Tim Burton" includes \textit{Tim Burton}, \textit{Tim Burton (musician)}, etc., and the list of ``directed by" includes \textit{director}.

The candidates in exhaustive lists are initially ranked according to text similarity, and irrelevant vertices/edges are usually included, e.g., \textit{Tim Burton (musician)} in the above list. Therefore, Dubey et al.~\cite{dubey2018earl} proposed two solutions to disambiguate the retrieved lists, including a formalization of Generalized Traveling Salesman Problem (GTSP) and a machine learning classifier. Since the GTSP-based solution can only select the optimal candidate of each list, which consequences a poor recall score, we employ the classifier-based solution in this paper. The classifier is designed based on a postulate: regarding one NLQ containing multiple entity/relation phrases, the correct mapping results of all the phrases tend to exhibit relatively dense and short-hop connections among themselves in the underlying KG compared to wrong results. The postulate has been validated~\cite{dubey2018earl}, and it is intuitive. For example, when we are trying to select the better candidate of ``Tim Burton" between \textit{Tim Burton} and \textit{Tim Burton (musician)}, we can compare the distances of \textit{Tim Burton} and \textit{Tim Burton (musician)} to the candidates of other entity/relation phrases of the given NLQ, e.g., \textit{director}. Apparently, \textit{Tim Burton} as a director has a smaller distance to other movie-related candidates compared to \textit{Tim Burton (musician)}. 

Dubey et al.~\cite{dubey2018earl} defined three features for the classifier: Text Similarity-based Initial Rank, Connection-Count, and Hop-Count. Text Similarity-based Initial Rank is computed during the retrieval of exhaustive lists. Connection-Count and Hop-Count are both computed based on the subdivision knowledge graph~\cite{dubey2018earl} to measure the connection situation of the candidate of one phrase to the candidates of all the other phrases. An extreme gradient boosting (xgboost)~\cite{chen2015xgboost} classifier is trained to compute the probability of a candidate being the most suitable one. Then, each exhaustive list is sorted according to the probabilities. For example, the sorted candidate list of ``Tim Burton" is \{\textit{Tim Burton}, \textit{Tim Burton Productions}, \textit{Burton}, etc.\}. 

\begin{figure*}
	\centering
	\includegraphics[width=0.55\linewidth]{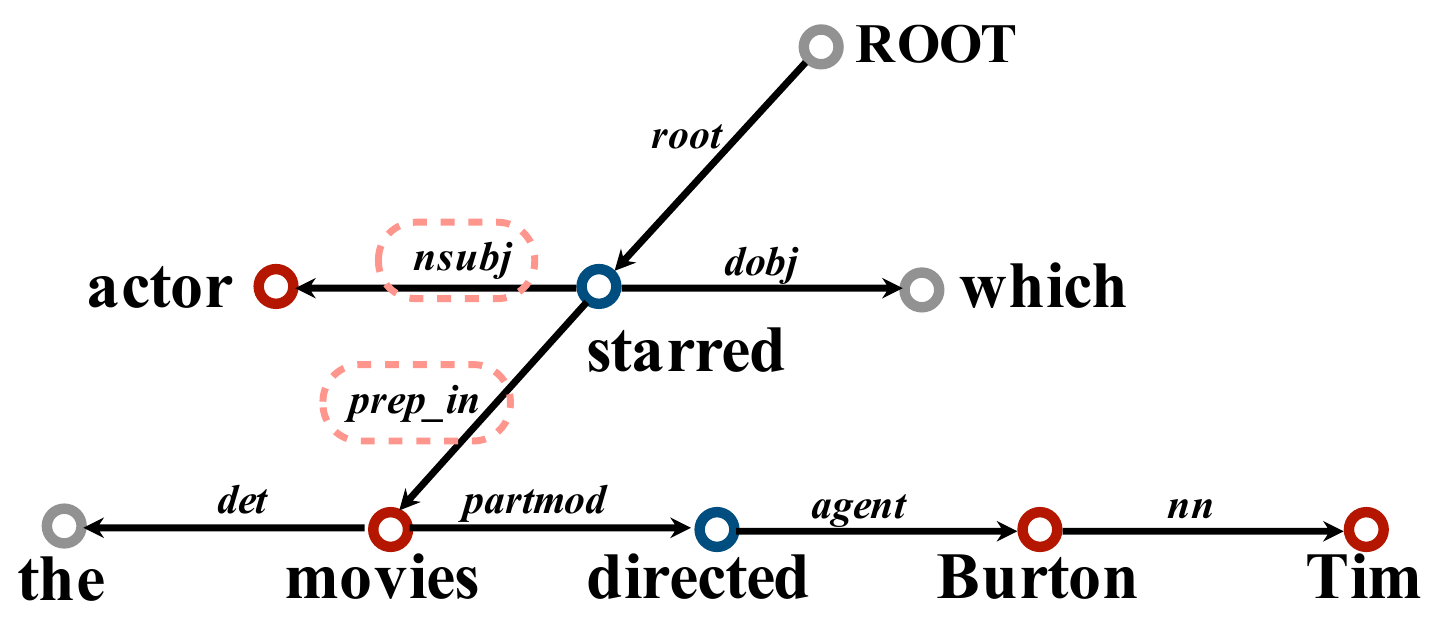}
	\caption{Dependency tree of the example NLQ.}
	\label{dependency-tree}
\end{figure*}

The above method can generate state-of-the-art phrase mapping results on question answering benchmarks~\cite{trivedi2017lc}. However, we still need to do the following refinements. Firstly, in practice, we found that the above method tends to classify extracted keywords into relation phrases. For example, when processing the NLQ ``\textit{which actor starred in the movies directed by Tim Burton}", both ``actor" and ``movies" are incorrectly classified into relation phrases. Inspired by~\cite{hu2018answering}, we utilize the dependency tree~\cite{de2008stanford} to address this issue. The dependency tree of an NLQ is a directed graph, where vertices represent words of the NLQ, and edges represent grammatical relations between words. The dependency tree of the example NLQ is illustrated in Fig.~\ref{dependency-tree}. Hu et al.~\cite{hu2018answering} summarized subject-like grammatical relations (e.g., \textit{subj}, \textit{nsubj}, and \textit{nsubjpass}) and object-like grammatical relations (e.g., \textit{obj}, \textit{pobj}, and \textit{dobj}) to extract associated entity phrases of the recognized relation phrases in the dependency tree. We do not need to specify the association relations between entity/relation phrases, but the grammatical relations can be utilized to check the initial entity/relation classification results of the above method. 

Specifically, given the initially classified entity/relation phrases, if a relation phrase is pointed by subject/object-like grammatical relations from other relation phrases, then we change its category to the entity phrase. In the dependency tree of the example NLQ, ``actor" is pointed by \textit{nsubj} from another relation phrase ``starred", then we know that ``actor" is actually an entity phrase, as illustrated in Fig.~\ref{dependency-tree}. We also change the categories of relation phrases which are pointed by prepositional grammatical relations (e.g., \textit{prep\_in} and \textit{prep\_of}). For example, ``movies" is pointed by \textit{prep\_in}, and we change the category of ``movies" to the entity phrase. It is worth mentioning that, if an initially classified relation phrase has entity phrase neighbors, or it has prepositional grammatical relations (e.g., \textit{prep\_in} and \textit{prep\_of}) pointing to other phrases, we do not further change its category. For example, in the dependency tree of the example NLQ, ``starred" has the out grammatical relation \textit{prep\_in}, and ``directed" has the entity phrase neighbor ``Tim Burton". Therefore, we assume that the initial classification results (i.e., relation phrases) of ``starred" and ``directed" are correct. The first reason is that if we change the category of an initially classified relation phrase which has an entity phrase neighbor, then we get a pair of entity phrases being directly linked in the dependency tree, which means that an implied semantic relation between the two entity phrases is manually created~\cite{hu2018answering}. The second reason is that phrases which have out prepositional grammatical relations are basically relation phrases in common NLQs, except the NLQs which use prepositions as relation phrases, e.g., ``\textit{list the schools in Germany}".

The following processes are also necessary to be performed on the mapping results. Firstly, the sorted lists of entity/relation phrases may contain both candidate vertices and edges at the same time. Therefore, we delete candidate vertices from the sorted lists of relation phrases, and vice versa. After the deletion, there may still exist too many candidates in the lists. Therefore, for each list, assuming that the probability value (computed by the classifier) of the first candidate is $p_v$, we set a threshold parameter $t_s$ and delete the candidates whose probability values are less than $p_v/t_s$. Another issue of the above mapping method is that wh-words (e.g., ``what", ``who", and ``where") are not processed. For example, given the NLQ ``\textit{who is the mayor of Berlin}", we can obtain the entity phrase ``Berlin" and the relation phrase ``mayor of". However, one more entity phrase is needed to construct a triple pattern in the following modules. Therefore, for NLQs which use wh-words to denote variables, we extract the wh-words as entity phrases and map them according to the DBpedia Ontology Class\footnote{http://mappings.dbpedia.org/server/ontology/classes/}. In this example, we map ``who" to \textit{Person} and \textit{Agent}. For NLQs which have implied entity phrases, we also add wh-words as intermediate entity phrases. For example, the NLQ ``\textit{who is the mayor of the capital of Germany}", can be paraphrased as ``\textit{who is the mayor of the city which is the capital of Germany}", and the entity phrase ``city" is implied. Besides the original entity phrases \{``who", ``Germany"\} and the original relation phrases \{``mayor of", ``capital of"\}, we add ``what" as an intermediate entity phrase since its matching vertex is \textit{Thing} which is the base class of all ontology classes. Finally, we recall the related annotations as follows: given an NLQ $\mathcal{Q}$, the extracted entity phrases and relation phrases are respectively denoted as $\{ent_1, ent_2, ..., ent_n\}$ and $\{rel_1, rel_2, ..., rel_m\}$. The candidate vertex set of $ent_i$ is denoted as $C_v^i$, and the candidate edge set of $rel_j$ is denoted as $C_e^j$.

\subsection{Structure Computing}
\label{Structure Computing}
In this module, we compute the optimal structure of the target query in the learned embedding space, where vertices/edges of the same candidate set are close to each other, as we analyzed in Section~\ref{KG Embedding Learning}. Firstly, each candidate vertex/edge set is regarded as an individual vertex/edge whose embedding representation is the mean embedding vector of the vertices/edges of the candidate set. Specifically, the embedding representations of candidate vertex set $C_v$ and candidate edge set $C_e$ are respectively computed as follows:

\begin{equation}
\mathbf{C_v}=\frac{1}{|C_v|}\sum_{v\in C_v}\mathbf{v}.
\end{equation}

\begin{equation}
\mathbf{C_e}=\frac{1}{|C_e|}\sum_{e\in C_e}\mathbf{e}.
\end{equation}

Then we assemble candidate vertex/edge sets into structure graphs to represent possible structures of the target query, as illustrated on the right side of Fig.~\ref{framework}(b). In the following of this paper, the structure graph is denoted by the structure matrix which is defined as follows:

\begin{defi}[Structure Matrix] \label{Structure matrix} The structure graph consisting of $n$ candidate vertex sets $\{C_v^1, C_v^2, ..., C_v^n\}$ and $m$ candidate edge sets $\{C_e^1, C_e^2, ..., C_e^m\}$ is denoted by the structure matrix $M_S$:
\[
M_S=	\left[ \begin{array}{cccc}
ms_{1,1} & ms_{1,2} & ... & ms_{1,n}\\
ms_{2,1} & ms_{2,2} & ... & ms_{2,n}\\
. & . & ... & .\\
. & . & ... & .\\
. & . & ... & .\\
ms_{n,1} & ms_{n,2} & ... & ms_{n,n}\\
\end{array} 
\right ].
\]

For each candidate vertex set $C_v^{i}$ in the structure graph, if $C_v^{i}$ is linked to another candidate vertex set $C_v^{j}$ by the candidate edge set $C_e^{k}$, then $ms_{i,j}=k$. If $C_v^i$ is not linked to $C_v^j$, then $ms_{i,j}=0$. If the structure matrix $M_S$ satisfies the following constraints, then its corresponding structure graph represents a valid structure of the target query.
\begin{enumerate}
	\item If $i=j$, $ms_{i,j}=0$;
	
	\item If $ms_{i,j} \neq 0$, $ms_{j,i}=0$;
	
	\item The number of non-zero elements in $M_S$ is $m$;
	
	\item For an integer $\alpha$, if $1 \leq \alpha \leq n$, $\sum_{i=1}^{n}ms_{i,\alpha}+\sum_{j=1}^{n}ms_{\alpha,j} \neq 0$;
	
	\item For an integer $\beta$, if $1\leq \beta \leq m$, there is an element $ms_{i,j}=\beta$ in $M_S$.
\end{enumerate}
\end{defi}

Intuitively, we can assemble candidate vertex/edge sets into a large set of structure graphs. However, not every assembled structure graph represents a valid structure of the target query. The above constraints are sufficient and necessary conditions for a structure graph to be valid. The first constraint means that in the structure graph, candidate vertex sets should not be linked to themselves. The second constraint means that for any two candidate vertex sets $C_v^{i}$ and $C_v^{j}$, if $C_v^{i}$ is linked to $C_v^{j}$, then $C_v^{j}$ should not be linked to $C_v^{i}$. The third constraint means that there should be $m$ candidate edge sets in the structure graph. The fourth constraint means that if we regard the structure graph as an undirected graph containing $n$ different vertices, the graph should be connected, i.e., there are no unreachable candidate vertex sets in the structure graph. The last constraint means that all the candidate edge sets should be assembled into the structure graph.

Taking the structure graph shown in Fig.~\ref{framework}(b) as an example, it can be denoted by the following structure matrix:
\[
M_S=	\left[ \begin{array}{cccc}
0 & 0 & 0\\
1 & 0 & 2\\
0 & 0 & 0\\
\end{array} 
\right ].
\]

The possibility of a structure graph representing the optimal structure of the target query is measured by the cost score $CS(\cdot)$ of its corresponding structure matrix $M_S$. The cost score is computed based on the translation mechanism, and a small cost score means a high possibility.
 \begin{equation}
 \label{structure evaluate function}
CS(M_S) = \sum_{\forall ms_{i,j}\neq0, ms_{i,j}\in M_S}\|\mathbf{C_v^i} + \mathbf{C_e^{ms_{i,j}}} - \mathbf{C_v^j}\|_2^2.
 \end{equation}
 
Then, the problem of deducing the optimal structure of the target query can be converted into the problem of finding the valid structure graph whose structure matrix has the minimum cost score. Given an NLQ containing $n$ entity phrases and $m$ relation phrases, the corresponding $n$ candidate vertex sets and $m$ candidate edge sets can be assembled into $n^{2m}$ possible structure graphs. It is time-consuming to generate all possible structure graphs, check whether they are valid, and compute their cost scores. Therefore, we propose Algorithm~\ref{Generating the optimal structure matrix.} to generate the structure graph whose structure matrix has the minimum cost score efficiently.
 
 The basic idea of Algorithm~\ref{Generating the optimal structure matrix.} is that we first generate an ideal structure graph in which the cost of assembling each candidate edge set is minimum (Line 1 to Line 12). Note that the generated ideal structure graph may be not valid. Specifically, for each candidate edge set $C_e^k, k = 1, ..., m$, we calculate the cost of linking any two candidate vertex sets by $C_e^k$ and store all possible costs in the two-dimensional array $Cost_k$. If the element $Cost_k[\alpha][\beta]$ has the minimum value in $Cost_k$, linking candidate vertex sets $C_v^\alpha$ to $C_v^\beta$ has the minimum cost for $C_e^k$, and we set $ms_{\alpha, \beta} = k$ in the initial matrix $M_S$. After performing this process for each candidate edge set, $M_S$ is the matrix which denotes the ideal structure graph. Then, if $M_S$ satisfies the above five constraints, the ideal structure graph is valid, and it represents the optimal structure of the target query (Line 13 to Line 14). However, due to possible errors of the learned embedding vectors, $M_S$ may do not satisfy the constraints, which means that there are candidate edge sets being incorrectly assembled. We first assume that only one candidate edge set is incorrectly assembled and try to modify $M_S$ into a valid structure matrix by changing the candidate vertex sets on the two sides of one candidate edge set every time. If we fail, then there are multiple candidate edge sets being incorrectly assembled, and we try to change the candidate vertex sets on the two sides of multiple candidate edge sets every time (Line 15 to Line 24). Specifically, in the function MODIFY(), we firstly select a set of candidate edge sets $\{C_e^k | k = K_1, ..., K_{num}, 1 \leq K_1, ..., K_{num} \leq m\}$ which are suspected to be incorrectly assembled (Line 26 to Line 36). Then, the function CHANGE() is called to change the candidate vertex sets on the two sides of each selected candidate edge set (Line 38 to Line 54).

\qquad
 
\begin{breakablealgorithm}
	\caption{Generating the structure matrix of the optimal structure graph.}
	\label{Generating the optimal structure matrix.}
	\begin{algorithmic}[1]
		\Require embedding vectors of vertices/edges of the underlying KG, $n$ candidate vertex sets $\{C_v^1, C_v^2, ..., C_v^n\}$, $m$ candidate edge sets $\{C_e^1, C_e^2, ..., C_e^m\}$, initial cost score $\hat{cost} = 100$, $n\times n$ matrices $\hat{M_S}$ and $M_S$, where $\hat{ms}_{i,j} = ms_{i,j} = 0$, $i = 1, ..., n$, $j = 1, ..., n$, an integer $num$.
		
		\Ensure the structure matrix $M_S$ which denotes the optimal structure graph.
		
		\State Create $m$ two-dimensional $n$-by-$n$ arrays $COST = \{Cost_k | k =1, ..., m\}$;
		\For{each array $Cost_k, k = 1, ..., m$}
			\For{each element $Cost_k[i][j], i = 1, ..., n, j = 1, ..., n$}
				\If{$i \neq j$}
					\State Set $Cost_k[i][j] = \|\mathbf{C_v^i} + \mathbf{C_e^{k}} - \mathbf{C_v^j}\|_2^2$;
				\Else
					\State Set $Cost_k[i][j] = \hat{cost}$;
				\EndIf
			\EndFor
			\State Find the element $Cost_k[\alpha][\beta]$ which has the minimum value in $Cost_k$;
			\State Set $ms_{\alpha, \beta} = k$;
		\EndFor
		\If{$M_S$ is a valid structure matrix}
			\State \Return{$M_S$}
		\Else
			\For{$num = 1, num \leq m, num$++}
				\State \Call {Modify}{$num, COST, M_S, \hat{M_S}, \hat{cost}$};
				\If{$\hat{cost} \neq 100$}
					\State Set $M_S = \hat{M_S}$;
					\State \Return{$M_S$}
				\EndIf
			\EndFor
		\EndIf
		\State \Return
		\State 
		\Function {Modify}{$num, COST, M_S, \hat{M_S}, \hat{cost}$}
			\For{each subset of $COST$ which contains $num$ arrays: $COST' = \{Cost_k | k = K_1, ..., K_{num}, 1 \leq K_1, ..., K_{num} \leq m\} \subseteq COST$}
				\State Create an $n\times n$ matrix $M_S'$ and set $M_S' = M_S$, i.e., $ms_{i, j}' = ms_{i, j}, i = 1, ..., n, j = 1, ..., n$;
				\For{each element $Cost_k \in COST'$}
					\State Find the element $ms_{\alpha, \beta}' = k$ in $M_S'$;
					\State Set $ms_{\alpha, \beta}' = 0$;
				\EndFor
				\State \Call {Change}{$num, COST', M_S', \hat{M_S}, \hat{cost}$}
			\EndFor
			\State \Return{}
		\EndFunction
		\State
		\Function {Change}{$num, COST', M_S', \hat{M_S}, \hat{cost}$}
			\For {each element $Cost_{K_{num}}[\alpha'][\beta']$ of $Cost_{K_{num}}\in COST'$, where $\alpha' = 1, ..., n, \beta'= 1, ..., n$}
				\State Create the integer $tmp = ms'_{\alpha', \beta'}$;
				\State Set $ms'_{\alpha', \beta'} = K_{num}$;
				\If {$num > 1$}
					\State \Call {Change}{$num-1, COST', M_S', \hat{M_S}, \hat{cost}$}
				\Else
					\If{$M_S'$ is a valid structure matrix}
						\If{$CS(M_S') < \hat{cost}$}
							\State Set $\hat{M_S} = M_S'$, $\hat{cost} = CS(M_S')$;
						\EndIf
					\EndIf
				\EndIf
				\State Set $ms'_{\alpha', \beta'} = tmp$;
			\EndFor
			\State \Return{}
		\EndFunction
	\end{algorithmic}
\end{breakablealgorithm}

\subsection{Query Generation}
\label{Query Generation}

\begin{figure*}
	\centering
	\includegraphics[width=0.99\linewidth]{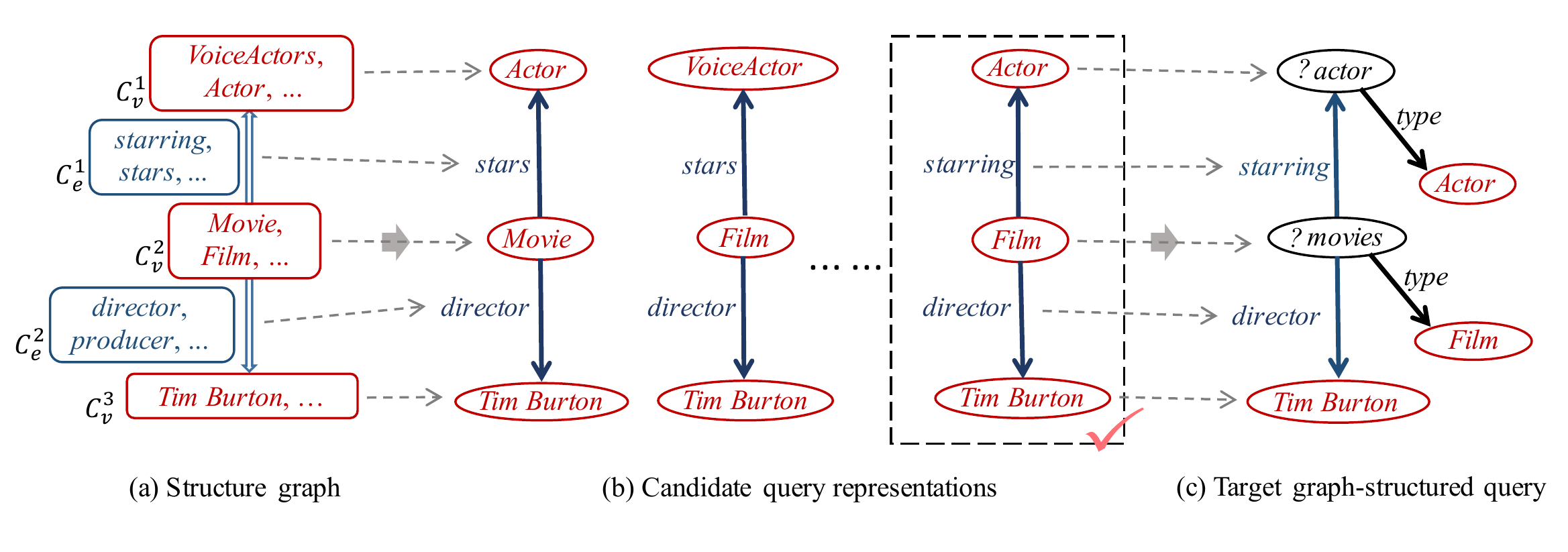}
	\caption{Generation of the target graph-structured query.}
	\label{queryGeneration}
\end{figure*}

In this module, we determine the most suitable matching vertex/edge of each entity/relation phrase and generate the target graph-structured query based on the computed optimal query structure, as illustrated in Fig.~\ref{queryGeneration}.

Taking the NLQ ``\textit{which actor starred in the movies directed by Tim Burton}" as an example, after the above two modules, we can obtain the optimal structure graph shown in Fig.~\ref{queryGeneration}(a). By selecting a vertex/edge from each candidate vertex/edge set, multiple candidate query representations can be constructed, as shown in Fig.~\ref{queryGeneration}(b). We denote the candidate query representation as a triple set $\mathcal{Q_R} = \{(v_h^1, e^1, v_t^1), (v_h^2, e^2, v_t^2), ..., (v_h^m, e^m, v_t^m)\}$ and evaluate $\mathcal{Q_R}$ by computing its cost score:
\begin{equation}
\label{query evaluate function}
Score(\mathcal{Q_R}) = \sum_{(v_h, e, v_t)\in \mathcal{Q_R}}\|\mathbf{v_h} + \mathbf{e} - \mathbf{v_t}\|_2^2.
\end{equation}
The candidate query representation $\mathcal{Q_R'}$ which has the minimum cost score is the optimal. We generate the target graph-structured query by replacing class vertices in $\mathcal{Q_R'}$ with variables constrained by the original class vertices, as illustrated in Fig.~\ref{queryGeneration}(c).

If the structure graph consists of $n$ candidate vertex sets $\{C_v^1, C_v^2, ..., C_v^n\}$ and $m$ candidate edge sets $\{C_e^1, C_e^2, ..., C_e^m\}$, the number of candidate query representations is $\prod_{i=1}^{n}\prod_{j=1}^{m}|C_v^i|\cdot|C_e^j|$. Since most real-world NLQs contain less than seven entity/relation phrases~\cite{Unger2016}, the values of $m$ and $n$ are very limited. Besides that, the evaluation of candidate query representations is performed in the learned embedding space through numerical calculation. Therefore, the above method for selecting the most suitable matching vertices/edges and generating the target graph-structured query is feasible.

\subsection{Time Complexity Analysis}
\label{Time Complexity Analysis}
In this section, we present a time complexity analysis of two algorithms adopted in our framework. Firstly, in the phrase mapping module, we employ the Connection Density algorithm proposed in \cite{dubey2018earl} to compute two input features of the disambiguation classifier, i.e., Connection-Count and Hop-Count. As we have introduced in Section~\ref{Phrase Mapping}, for a candidate vertex/edge which corresponds to an entity/relation phrase of the given NLQ, its Connection-Count and Hop-Count are computed based on the hop distances between itself and all candidate vertices/edges of the other entity/relation phrases. Therefore, the elementary operation of the Connection Density algorithm is to compute the hop distance between two objects, which can be vertices or edges, in the subdivision knowledge graph~\cite{dubey2018earl}. Assuming that the given NLQ contains $L$ entity/relation phrases, and each phrase corresponds to $N$ candidates, then the distances between $N^2{{L}\choose{2}}$ pairs of objects need to be computed. Since ${{L}\choose{2}} \leq L^2$, the time complexity of the Connection-Density algorithm is $\mathcal{O}(N^2L^2)$.

The second algorithm we employed is Algorithm~\ref{Generating the optimal structure matrix.}, which generates the structure matrix of the optimal structure graph. In Algorithm~\ref{Generating the optimal structure matrix.}, we first construct the ideal structure graph whose cost score is minimum. Then, if the ideal structure graph is not valid, we try to re-assemble candidate edge sets to obtain a valid structure graph. In the worst case, all candidate edge sets need to be re-assembled, and the elementary operation is to check whether the intermediately modified structure graph is valid. Assuming that there are $n$ candidate vertex sets and $m$ candidate edge sets, the check operation would be performed for $\sum_{k=0}^{m}{{m}\choose{k}}(n^2)^k$ times. Since ${{m}\choose{k}} \leq m^k$, $\sum_{k=0}^{m}{{m}\choose{k}}(n^2)^k \leq \sum_{k=0}^{m}(mn^2)^k$, and $\sum_{k=0}^{m}(mn^2)^k = \frac{1-(mn^2)^{m+1}}{1-mn^2}$, the time complexity of Algorithm~\ref{Generating the optimal structure matrix.} is $\mathcal{O}(m^mn^{2m})$. It is worth mentioning that Algorithm~\ref{Generating the optimal structure matrix.} actually stops once a valid structure graph is constructed, and the number of entity/relation phrases of real-world NLQs, i.e., $m$ and $n$, is very limited. Therefore, Algorithm~\ref{Generating the optimal structure matrix.} is feasible, and it has been validated in our experiment.

\subsection{Discussion}
\label{Discussion}
We discuss three issues which need to be considered during the framework implementation: 1) It is a common scenario where relation phrases are implied in the NLQ. For example, the NLQ ``\textit{List actors born in Germany.}" is usually expressed as ``\textit{List German actors}", where the relation phrase ``born in" is implied. We employ the method in~\cite{hu2018answering} to generate candidate edges for implied relation phrases. 2) If the graph-structured query generated by the optimal query representation returns an empty answer, and the problem cannot be addressed deleting the constraints of class vertices on variables, we would generate another query based on candidate query representations with higher cost scores. 3) Other improved translation-based models such as TransH~\cite{wang2014knowledge} and TransR~\cite{lin2015learning} can also be adopted in our framework by modifying the functions computing cost scores, e.g., Equ.~\ref{vertex score function} and Equ.~\ref{edge score function}. The performance may be improved by adopting improved translation mechanisms. However, the framework would be more complicated at the same time, and the training time would increase rapidly. We will conduct a more in-depth investigation into this part in the future. 

\section{Experiments}
\label{Experiments}
The graph-structured queries constructed by our framework can be evaluated over KGs to obtain the answers of given NLQs. To scrutinize the effectiveness and efficiency of our framework, we compare it with KG-based question answering models, including all models participating in the first task of QALD-6~\cite{Unger2016} and two state-of-the-art models RFF~\cite{hu2018answering} and NFF~\cite{hu2018answering}. We also validate our embedding method by comparing it with TransE in terms of the translation mechanism and providing a visualization of sample learned embedding vectors. All experiments were conducted on a Linux server with an Intel Core i7 3.40GHz CPU and 128GB memory running Ubuntu-14.04.1, and we set the dimension of embedding vectors to 100, $\lambda_v = 0.5$, $\lambda_e = 0.5$, and $t_s = 15$.

\subsection{Dataset}
\label{Dataset}
\noindent \textbf{KG Dataset.} DBpedia is a large-scale KG which contains structured information extracted from Wikipedia\footnote{\url{https://www.wikipedia.org/}}. We employ the version of DBpedia-2015\footnote{\url{http://wiki.dbpedia.org/develop/datasets}} which consists of 6.7M vertices, 1.4K edges, and 583M KG triples.

\noindent \textbf{NLQ Dataset.} QALD-6~\cite{Unger2016} is the sixth installment of a series of challenges on question answering over KGs.\footnote{The later published installment contains a large part of NLQs demanding mathematical operations and questions according to RDF types~\cite{usbeck20177th}, which are not the focus of this paper.} It published 100 test questions over DBpedia for the first task ``Multilingual Question Answering"~\cite{Unger2016}. And the test questions are associated with gold graph-structured queries and answers.

\subsection{Effectiveness Evaluation}
\label{Effectiveness Evaluation}

In this section, we follow~\cite{Unger2016} to evaluate the effectiveness of our framework with three metrics: recall, precision, and F-1 measure. Recall refers to the ratio of correct answers obtained by the constructed query over all gold answers. Among all answers obtained by the constructed query, precision refers to the proportion of correct answers. F-1 measure is a weighted average between precision and recall, and it is computed as follows:
\begin{equation}
F-1\ measure = \frac{2\cdot Precision\cdot Recall}{Precision+Recall}.
\end{equation}

\begin{table}
	\caption{Results on the NLQ Dataset (Total number of questions: 100)}
	\label{effectiveness-result}
	\begin{tabular}{lllll}
		\noalign{\smallskip}\cline{1-5}\noalign{\smallskip}
		& Processed & Recall & Precision & F-1  \\ \noalign{\smallskip}\cline{1-5}\noalign{\smallskip}
		CANaLI~\cite{mazzeo2016answering}              & 100       & 0.89   & 0.89      & 0.89 \\ \noalign{\smallskip}
		\textbf{Our framework}        & \textbf{100}       & \textbf{0.73}     & \textbf{0.85}        & \textbf{0.79}   \\ \noalign{\smallskip} 
		NFF~\cite{hu2018answering}                 & 100       & 0.70   & 0.89      & 0.78 \\ \noalign{\smallskip}
		UTQA~\cite{veyseh2016cross}                & 100       & 0.69   & 0.82      & 0.75 \\ \noalign{\smallskip}
		KWGAnswer~\cite{hu2018answering}           & 100       & 0.59   & 0.85      & 0.70 \\ \noalign{\smallskip}
		RFF~\cite{hu2018answering}                 & 100       & 0.43   & 0.77      & 0.55 \\ \noalign{\smallskip}
		NbFramework~\cite{Unger2016}         & \textbf{63}        & 0.85   & 0.87      & 0.54$^*$ \\ \noalign{\smallskip}
		SemGraphQA~\cite{Unger2016}          & 100       & 0.25   & 0.70      & 0.37 \\ \noalign{\smallskip}
		UIQA(with manual)~\cite{Unger2016}   & \textbf{44}        & 0.63   & 0.54      & 0.25 \\ \noalign{\smallskip}
		UIQA(without manual)~\cite{Unger2016} & \textbf{36}        & 0.53   & 0.43      & 0.17 \\ \noalign{\smallskip}\cline{1-5}
	\end{tabular}\\
\footnotesize{(*Although NbFramework has quite high precision and recall, its F-1 measure is \\very low since F-1 measure is computed with respect to the total number of NLQs.)}\\
\end{table}

We report the evaluation results of our framework, RFF, NFF, and the models participating QALD-6 in Table \ref{effectiveness-result}, where \textit{Processed} indicates the number of processed NLQs for each model. Note that the recall and precision of each model are computed with respect to the number of processed NLQs, and F-1 measure is computed with respect to the total number of questions.

We sum up three observations based on Table \ref{effectiveness-result}: 

\begin{enumerate}
	
	\item Our framework is ranked second according to the F-1 measure. Nevertheless, our framework is still the most competitive one since the first-ranked system CANaLI~\cite{mazzeo2016answering} can only answer NLQs expressed by the Controlled Natural Language~\cite{mazzeo2016answering}.
	
	\item Among the models which processed all test NLQs, our framework achieves the highest recall except CANaLI, which is due to the following reasons: Firstly, the semantic gap between NLQs and KGs is well addressed by our framework since the query structure is computed based on the learned embedding vectors which are essentially the latent representation of the underlying KG. Secondly, our framework does not need to prune the candidate sets before the structure deducing and query generation, and the most suitable vertices/edges are selected during the query generation. 3) The selected matching vertices/edges of our framework are globally optimal since the selection is based on the cost score of the entire query.
	
	\item Among the models which processed all test NLQs, our framework is ranked second according to the precision except CANaLI. The main reason is that there are errors in the phrase mapping results generated by existing phrase mapping models~\cite{dubey2018earl, hu2018answering}, which cause incorrectly constructed queries.
	
\end{enumerate}

\subsection{Efficiency Evaluation}
\label{Efficiency Evaluation}
\begin{figure*}
	\centering
	\includegraphics[width=0.9\linewidth]{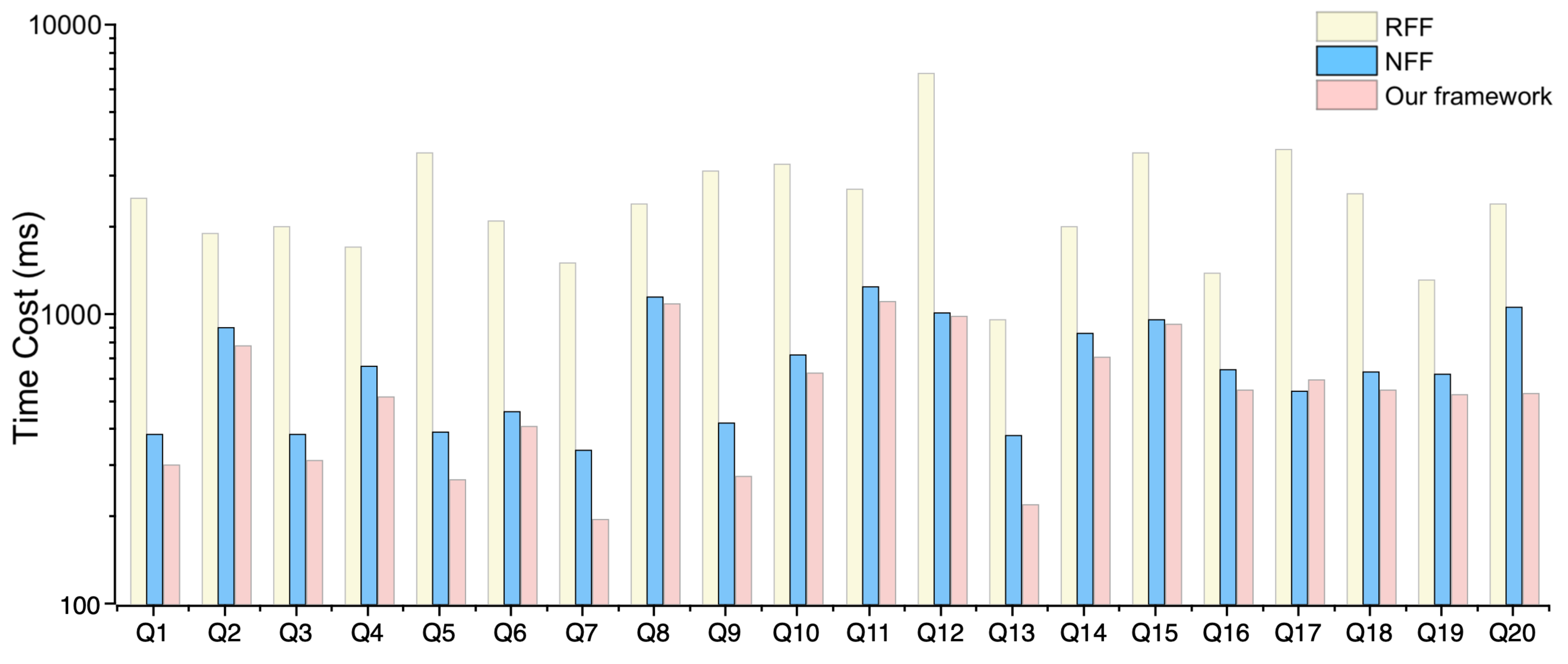}
	\caption{Time costs of our framework, RFF, and NFF.}
	\label{time_cost_comparison}
\end{figure*}

\begin{table}
	\caption{Average time cost of each module}
	\label{Average time cost of each phase}
	\begin{tabular}{ll}
		\noalign{\smallskip}\cline{1-2}\noalign{\smallskip}
		Module                & Avg.Time Cost (ms) \\ \noalign{\smallskip}\cline{1-2}\noalign{\smallskip}
		Phrase Mapping         & 475.1              \\ \noalign{\smallskip}
		Structure Computing   & 21.5              \\ \noalign{\smallskip}
		Query Generation and Evaluation & 104.8              \\ \noalign{\smallskip}\cline{1-2}
	\end{tabular}
\end{table}

The average time cost of our framework to construct the graph-structured query and obtain answers for a given NLQ is 601.4ms. The average time cost of each module is reported in Table \ref{Average time cost of each phase}. The phrase mapping module spends much more time than the other modules due to the large scale of DBpedia which also increases the time cost of the query evaluation process.

We randomly select 20 NLQs from the QALD-6 dataset. Time costs of our framework, RFF, and NFF to answer the twenty NLQs are illustrated in Fig.~\ref{time_cost_comparison}. Obviously, the time cost of our framework is less than the two baselines. This is reasonable since our framework leverages the learned embedding vectors to construct graph-structured queries. Both the query structure deducing and the selection of matching vertices/edges are performed by numerical calculations in the embedding space. In addition, the evaluation of constructed graph-structured queries is more efficient than the subgraph matching employed by RFF and NFF.

\subsection{Failure Analysis}
\label{Failure Analysis}

\begin{table}
	\caption{Failure analysis of our framework on the NLQ dataset}
	\label{Failure-analysis-table}
	\begin{tabular}{lll}
		\noalign{\smallskip}\cline{1-3}\noalign{\smallskip}
		Failure Module                                                        & \# (Ratio) & Sample Failure NLQ                                                                            \\ \noalign{\smallskip}\cline{1-3}\noalign{\smallskip}
		\begin{tabular}[c]{@{}l@{}}Phrase Mapping\end{tabular} & 13 (43.4\%)         & \begin{tabular}[c]{@{}l@{}}\textit{Which space probes were sent} \\\textit{into orbit around the sun?}\end{tabular} \\\noalign{\smallskip}
		Structure Computing                                                      & 4 (13.3\%)         & \begin{tabular}[c]{@{}l@{}}\textit{Who was the doctoral supervi-}\\\textit{sor of Albert Einstein?}\end{tabular}          \\\noalign{\smallskip}
		Query Generation                                                    & 3 (10.0\%)         & \begin{tabular}[c]{@{}l@{}}\textit{Who was Vincent van Gogh}\\ \textit{inspired by?}\end{tabular}            \\\noalign{\smallskip}
		Other                                                               & 10 (33.3\%)         & \begin{tabular}[c]{@{}l@{}}\textit{Give me a list of all critically}\\ \textit{endangered birds.}\end{tabular}           \\ \noalign{\smallskip} \cline{1-3}
	\end{tabular}
\end{table}

In this section, we analyze the failure causes of our framework. Given an NLQ, if the graph-structured query constructed by our framework cannot retrieve all the gold answers, or unrelated answers are retrieved, we consider that the framework cannot answer this NLQ correctly. Among the 100 test NLQs, our framework can correctly answer 70 NLQs. For the rest 30 NLQs, we divide them into four categories according to which module of the framework should be responsible for the failure. The analysis result is summarized in Table \ref{Failure-analysis-table}.

We can observe that the phrase mapping module should be responsible for most failures. There are mainly two reasons: firstly, some core entity/relation phrases of the given NLQ are implied or over-expressed. These phrases cannot be extracted by the phrase mapping module. For example, it is hard to identify correct relation phrases of the NLQ ``\textit{which space probes were sent into orbit around the sun}". Secondly, the phrase mapping module cannot obtain correct matching vertices/edges of some ambiguous entity/relation phrases. For example, given the NLQ ``\textit{what are the five boroughs of New York}", the phrase mapping module failed to map the relation phrase ``five boroughs of" to the edge \textit{governmentType}.

The structure computing module is responsible for four failures. The reason is that some candidate vertex/edge sets contain candidates whose embedding vectors are not close to each other. For example, given the NLQ ``\textit{who was the doctoral supervisor of Albert Einstein}", the candidate edge set of ``doctoral supervisor" contains two candidate edges: \textit{doctoralAdvisor} and \textit{doctoralStudent}. Since the learned embedding vectors of the two edges are not close to each other, the mean embedding vector of the candidate edge set cannot be used to compute the optimal query structure. 

Errors in the query generation module lead to three failures. The reason is that the computed cost score of the target query representation is not the minimum. For example, given the NLQ ``\textit{who was Vincent van Gogh inspired by}", the target query representation is $\mathcal{Q_R}=$ \{(\textit{Person}, \textit{influenced}, \textit{Vincent van Gogh})\}. However, the cost score of another query representation $\mathcal{Q_R'}=$ \{(\textit{Person}, \textit{influencedBy}, \textit{Vincent van Gogh})\} is lower, and $\mathcal{Q_R'}$ is incorrectly selected as the optimal query representation. The reason for this error is that, in the underlying KG, the number of people influenced by Vincent van Gogh is larger than the number of people influenced Vincent van Gogh. Therefore, during embedding learning, the triple in $\mathcal{Q_R'}$ attracts more attention according to the attention score computed by Equ.~\ref{attention-score-entity}. 

Ten NLQs cannot be correctly answered due to the limitation of our framework. The main reason is the requirement of operators which cannot be heuristically identified. For example, a UNION operator is required by the NLQ ``\textit{Give me a list of all critically endangered birds}". However, we cannot identify this requirement based on the NLQ itself.

\subsection{Embedding Method Validation}
\label{Embedding Method Validation}

The above evaluation of our framework has indirectly proved the effectiveness of our embedding method. For better comprehension and scrutiny, we conduct two additional experiments in this section to further validate the embedding method.

\begin{figure*}
	\centering
	\includegraphics[width=0.9\linewidth]{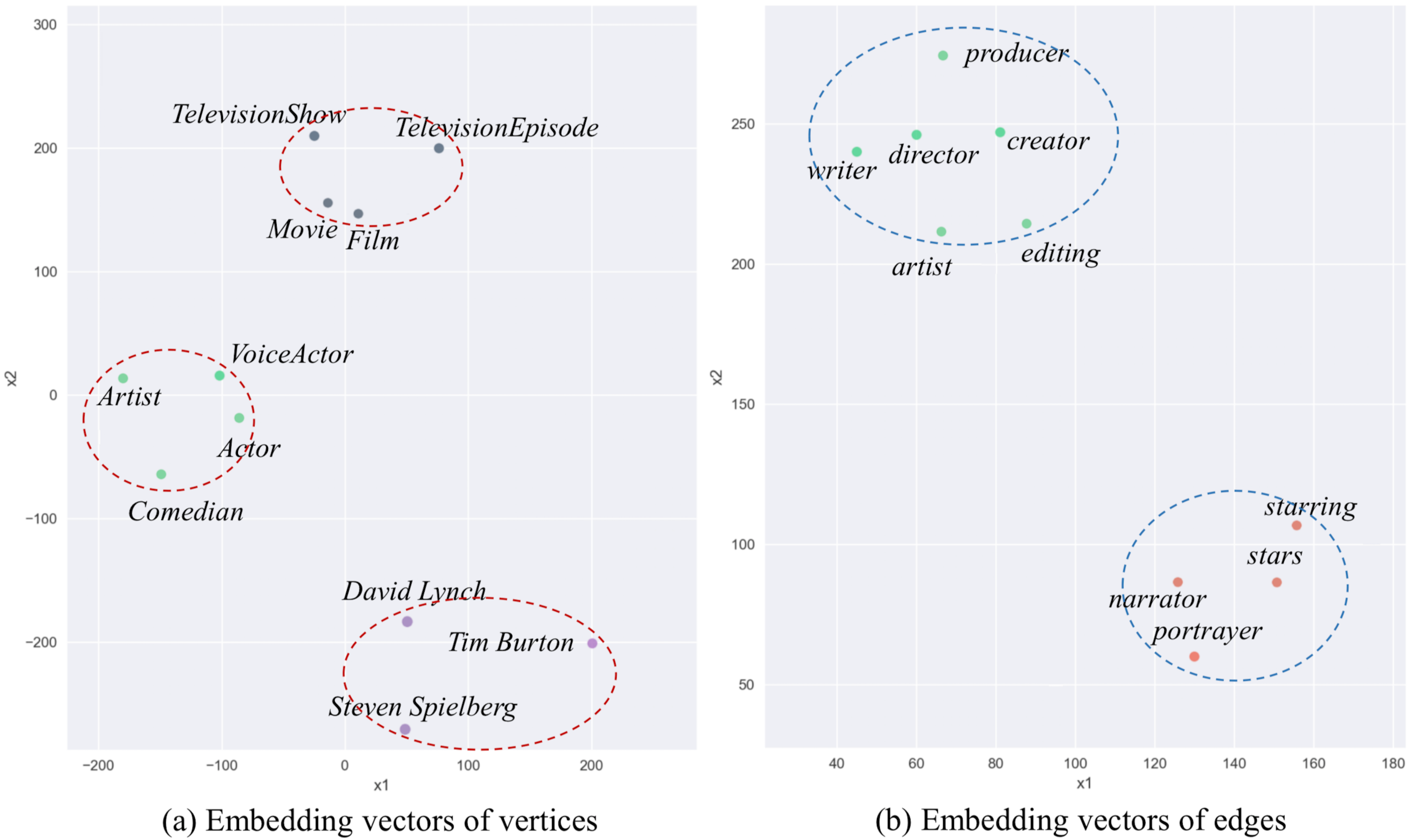}
	\caption{Visualization of learned embedding vectors.}
	\label{visual}
\end{figure*}

Firstly, we project sample learned embedding vectors into two-dimensional spaces using t-SNE\footnote{https://lvdmaaten.github.io/tsne/}. Sample embedding vectors of vertices and edges are respectively illustrated in Fig.~\ref{visual}(a) and Fig.~\ref{visual}(b). We can observe that, in the embedding space, semantically similar vertices/edges which share common GL-KGs are close to each other.

In terms of the translation mechanism, we compare our embedding method with TransE. Following the evaluation protocol used in~\cite{bordes2013translating}, we use \textit{MeanRank} and \textit{Hits@$10$} as evaluation metrics and employ FB15k~\cite{bordes2013translating} as the benchmark dataset. It is worth mentioning that in this evaluation, there is no need to generate GL-KGs for the vertices/edges in FB15k, and the embedding vectors are learned based on L-KGs. For each test KG triple ($v_h, e, v_t$), we first remove the head vertex $v_h$ or the tail vertex $v_t$. Then, we predict the missing vertex $v_h$ or $v_t$ based on $\mathbf{v_t} - \mathbf{e}$ or $\mathbf{v_h} + \mathbf{e}$ and use the cost function of TransE~\cite{bordes2013translating} to rank the predictions in a descending order. \textit{MeanRank} denotes the average rank of all correct predictions, and \textit{Hits@10} denotes the proportion of correct predictions ranked in top-$\textit{10}$. The \textit{MeanRank} of our embedding model is 261 and \textit{Hits@10} is 33.2. The \textit{MeanRank} of TransE is 243 and \textit{Hits@10} is 34.9. Both the results of \textit{MeanRank} and \textit{Hits@10} are very close, which proves that our embedding model maintains the translation mechanism of TransE effectively.

\section{Related Work}
\label{Related_Work}
In this section, we discuss related researches on the graph-structured query construction problem. Recently, a variety of techniques, including semantic parsing~\cite{zou2014natural,hu2018answering,berant2014semantic,hakimov2015applying,marginean2017question}, templates~\cite{bast2015more,unger2012template,zheng2015build,unger2012template}, the interaction with users~\cite{zheng2017natural,mazzeo2016answering}, and KG embeddings~\cite{bordes2014question,bordes2014open,han2017keyword,yang2014joint}, have been leveraged by query construction and question answering models.

Given an NLQ, Berant et al.~\cite{berant2014semantic} first use a deterministic procedure to construct multiple candidate queries, each of which is associated with heuristically generated natural language canonical utterances. Then, the optimal candidate query is selected according to paraphrasing scores of the associated canonical utterances with respect to the NLQ. Hu et al.~\cite{hu2018answering} first extract semantic relations of the NLQ from the corresponding dependency tree and construct semantic query graphs which represent the query intention. Then, semantic query graphs are matched in the underlying KG to find the target subgraph which contains answers. Query intention understanding and representation, e.g., the generation and scoring of canonical utterances in~\cite{berant2014semantic} and the semantic query graph construction in~\cite{hu2018answering}, lie at the core of semantic parsing-based models. These models achieve robust performances on ambiguous and expressive NLQs. For example, existing query construction models can easily answer the NLQ ``\textit{what is the profession of Tim Burton}". However, except semantic parsing-based models, most of them cannot answer the semantically similar NLQ ``\textit{what does Tim Burton do for a living}". With the focus on parsing NLQs, the information of underlying KGs, e.g., the schema and ontology, is usually ignored by semantic parsing-based models. Therefore, a weakness of semantic parsing-based models is that they cannot handle the ``semantic gap" between NLQs and KGs, as we analyzed in Section~\ref{Introduction}.

Unger et al.~\cite{unger2012template} first mirror the internal structure of the given NLQ by a graph-structured query template. Then, the template is instantiated by statistical entity identification and predicate detection. The main strength of template-based models is that, based on the underlying template set, they can directly deduce the structure of the target query according to the syntactic structure and occurring expressions~\cite{unger2012template} of the given NLQ. With the target structure known, phrase mapping and query generation processes would be very efficient. Another strength is that template-based models do not need heuristic rules for NLQs involving additional operations, e.g., comparison, counting, and intersection. For example, given the NLQ ``\textit{who directed the most movies}", our model needs to first generate the graph-structured query and then heuristically add an \textit{ORDER BY DESC()} operator according to the modifier ``most". However, template-based models can directly generate the target query with required operators according to the underlying matching template. The main weakness of template-based models is that they cannot process NLQs which do not have existing matching templates. Therefore, performances of template-based models heavily depend on the underlying template set. However, the existing template sets are still far from being full-fledged.

Zheng et al.~\cite{zheng2017natural} proposed an interaction-based query construction model following the similar pathway of this paper, i.e., phrase extraction, phrase mapping, and query assembly. The main contribution of the model in~\cite{zheng2017natural} is that it allows users to verify ambiguities during the query construction. For example, users can select the correct mapping results of ambiguous phrases. Interaction-based models can perform highly accurate disambiguation based on feedback from users. And the effectiveness of the query assembly process would also be improved based on accurate mapping results. However, the time cost for answering an NLQ is largely increased to more than twenty seconds~\cite{zheng2017natural}. Since the query construction is an online task, the user experience would be degraded.

Models proposed in~\cite{bordes2014question,bordes2014open} utilize embedding techniques to directly obtain answers without the query construction. They first encode the word vocabulary of NLQs and vertices/edges of the underlying KG into embedding spaces. Then, the candidate answer is evaluated based on the embedding representations of the answer itself and the given NLQ. Since the two models do not need to perform relation phrase mapping and disambiguation, frequent accesses to the underlying KG can be avoided. In addition, the generation of answers is performed in the embedding space by numeral calculations. Therefore, the two models are competitive in efficiency and can be applied to large-scale KGs. However, a relative large-scale training dataset of NLQs is required, and the performance relies on the training dataset heavily. Another weakness of the two models is that they do not consider the internal structure of the given NLQ, e.g., dependency relations and the syntactic structure. However, the internal structure is vital for answering complex NLQs containing multiple entity/relation phrases.

Han et al.~\cite{han2017keyword} proposed an embedding-based model to construct the graph-structured query of given keywords. The model first performs phrase classification and mapping and then assemble mapping results into candidate queries. Embedding representations of the underlying KG are utilized to speed up the evaluation of candidate queries, which makes the model applicable to large-scale KGs. A weakness of the model is that it directly constructs candidate queries without deducing the target structure, which limits the efficiency of the model. In addition, the leveraged embedding representations are directly learned by TransE without any further improvement. Therefore, the relations between class vertices and entity vertices cannot be well captured in the embedding space, which may cause errors during the candidate query evaluation.

Compared to the above embedding-based models, our framework has the following strengths: firstly, our framework generates candidate queries based on the deduced query structure, and the employed embedding representations are learned based on GL-KGs which contain the information related to class vertices. Therefore, the above two weaknesses of the model in~\cite{han2017keyword} are avoided. Secondly, different from~\cite{bordes2014question,bordes2014open}, our framework does not need the training dataset of NLQs, and the internal structure of the given NLQ is considered during the phrase mapping process. Therefore, our framework is able to process complex NLQs containing multiple entity/relation phrases. Essentially, we expect that the given NLQ is a faithful expression of its target graph-structured query in natural language, which consists of entity/relation phrases describing the vertices/edges of the target query. Therefore, unlike semantic parsing-based models, our framework cannot process expressive NLQs in which core entity/relation phrases are implied, and misleading or redundant phrases are mentioned, e.g., ``\textit{what does Tim Burton do for a living}". Another weakness of our framework is the propagation of errors along the three modules. As we analyzed in Section \ref{Failure Analysis}, most failures are caused by propagated errors from the phrase mapping module.

\section{Conclusion and Future Work}
\label{Conclusion}
In this paper, we propose a novel framework which leverages recent embedding techniques to construct graph-structured queries of given NLQs. Before the query construction, we first learn embedding representations of the underlying KG based on the GL-KGs of vertices/edges. Our embedding method maintains the translation mechanism and is able to capture the relations relevant to class vertices in the KG. In addition, vertices/edges sharing common GL-KGs are close to each other in the embedding space. Based on the learned embedding vectors, we represent the phrase mapping result of each entity/relation phrase as the mean embedding vector and propose an efficient algorithm to compute the optimal structure of the target query. Then, according to the computed query structure, we select the most suitable matching vertex/edge of each entity/relation phrase and generate the target query by adopting the translation mechanism. Extensive experiments have been conducted on the benchmark dataset. The results demonstrate that our framework outperforms other baseline models in terms of effectiveness and efficiency. 

Since the main failure cause of our framework is the errors during phrase mapping, we intend to explore more effective phrase mapping methods in the future. And, we are trying to improve the performance of our embedding method by adopting improved translation mechanisms.

\section*{Acknowledgment}
This work is supported by National Key Research and Development Program of China (No. 2018YFB1004500), National Natural Science Foundation of China (61532015, 61532004, 61672419, and 61672418), Innovative Research Group of the National Natural Science Foundation of China (61721002), Innovation Research Team of Ministry of Education (IRT\_17R86), Project of China Knowledge Centre for Engineering Science and Technology, Science and Technology Planning Project of Guangdong Province (No. 2017A010101029), Teaching Reform Project of XJTU (No. 17ZX044), and China Scholarship Council (No. 201806280450). We would like to express our gratitude to Mr. Zhouguo Chen for his advice during paper writing and experiments. The current work is an extension and continuation of our previous work that has been published in a conference paper of ICBK 2018~\cite{wang2018graph}.

\bibliographystyle{spbasic}
\bibliography{reference}

\section*{Author Biographies}
\leavevmode
\vbox{%
	\setlength\intextsep{0pt}
	\begin{wrapfigure}{l}{80pt}
			\includegraphics[width=1\linewidth]{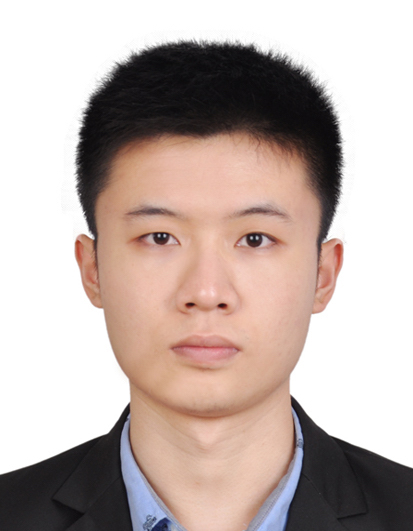}
	\end{wrapfigure}
	\noindent\small 
	{\bf Ruijie Wang} is currently working toward the MS degree in Computer Science at Xi'an Jiaotong University, Xi'an, China. He received the BS degree in Computer Science and the BS degree in Economics from Xi'an Jiaotong University in 2017. From 2018 to 2019, he was a joint master student at Informatik 5, RWTH Aachen University, Aachen, Germany. His research interests include semantic web, knowledge graph embedding, and question answering.}

\vspace{60pt}

\vbox{%
	\setlength\intextsep{0pt}
	\begin{wrapfigure}{l}{80pt}
				\includegraphics[width=1\linewidth]{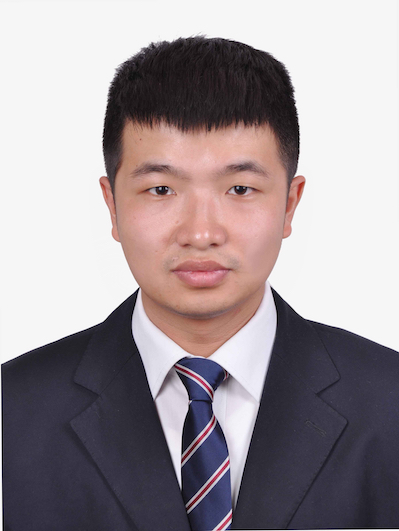}
	\end{wrapfigure}
	\noindent\small {\bf Meng Wang } is currently an assistant professor at the School of Computer Science and Engineering, Southeast University, Nanjing, China. He received the Ph.D. degree in computer science from Xi'an Jiaotong University in 2018 and BS degree in Computing Science from Sichuan University in 2012. His research interests include knowledge graph, semantic web, and data mining.}

\vspace{60pt}

\vbox{%
	\setlength\intextsep{0pt}
	\begin{wrapfigure}{l}{80pt}
		\includegraphics[width=1\linewidth]{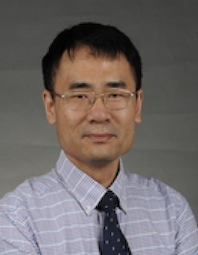}
	\end{wrapfigure}
	\noindent\small 
	{\bf Jun Liu} is currently a professor at the Department of Computer Science, Xi’an Jiaotong University, Xi'an, China. Prof. Liu is also a professor at Guang Dong Xi'an Jiaotong University Academy, Shunde, China. He has published more than 70 research papers in various journals and conference proceedings. He received the BS, MS, and Ph.D. degrees from Xi’an Jiaotong University in 1995, 1998 and 2004, respectively. His main research interests include text mining, data mining, and e-learning.}

\vspace{45pt}

\vbox{%
	\setlength\intextsep{0pt}
	\begin{wrapfigure}{l}{80pt}
		\includegraphics[width=1\linewidth]{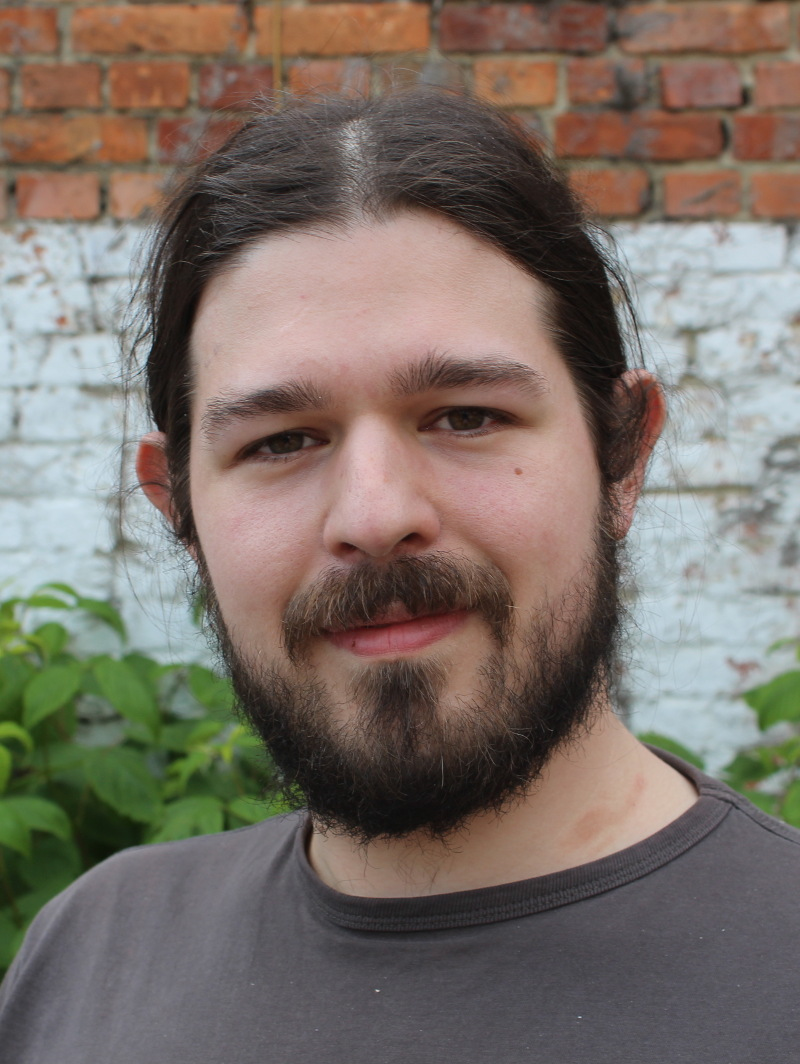}
	\end{wrapfigure}
	\noindent\small 
	{\bf Michael Cochez} is an assistant professor at VU Amsterdam, the Netherlands. Earlier he was a postdoctoral researcher at the Fraunhofer Institute for Applied Information Technology (FIT) and based at RWTH Aachen University, Germany. He received his Ph.D. and M.Sc. degrees in Mathematical Information Technology from University of Jyväskylä, Finland, and the B.Sc. degree in Information Technology from University of Antwerp, Belgium. His research interests include data analysis and knowledge representation, e.g., Knowledge Graph Embedding, Scalable Hierarchical Clustering, Prototype-based Ontologies, and Machine Learning}

\vspace{30pt}

\vbox{%
	\setlength\intextsep{0pt}
	\begin{wrapfigure}{l}{80pt}
		\includegraphics[width=1\linewidth]{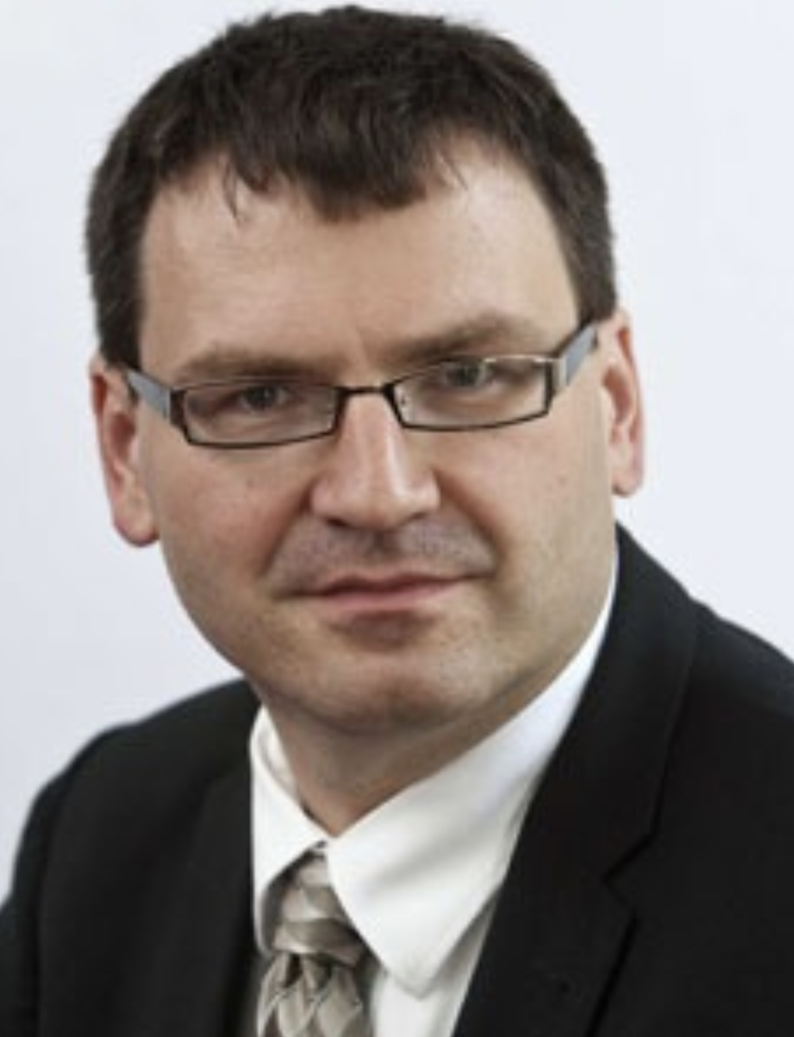}
	\end{wrapfigure}
	\noindent\small 
	{\bf Stefan Decker} is the director of Fraunhofer FIT and Full Professor of Information Systems and Databases at RWTH Aachen University, Germany. Earlier he served as Professor of Digital Enterprise at the National University of Ireland, and as Executive Director of the Digital Enterprise Research Institute (DERI). Prof. Decker studied Computer Science at the University of Kaiserslautern, completed his doctorate at the Business Faculty of Karlsruhe Technical University and did post-doctoral research work at Stanford University and the University of Southern California. His research interests include Semantic Web and linked data, knowledge representation, and data management.}\vspace{60pt}

\end{document}